%% file: main.tex
\ifdefined\pdfminorversion\pdfminorversion=7\fi
\documentclass[onecolumn]{main}
\ifdefined\paperstyle
  \paperstyle{simple}
  \papercolor{navy}
\fi
\IfFileExists{lmodern.sty}{}{\microtypesetup{expansion=false}}

\usepackage{xurl}
\usepackage{amsthm}
\usepackage{float}
\usepackage{threeparttable}
\usepackage{listings}
\lstdefinelanguage{yaml}{
  keywords={true,false,null},
  sensitive=false,
  comment=[l]{\#},
  morestring=[b]',
  morestring=[b]",
}
\hypersetup{pdfauthor={Zan Gong, Zengle Ge, Yui Lo, Linyue Pan, Bihui Yin, Guangxi Wan, Jinduo Zhang, Zhirong Zhang, Mo Zhao, Annan Li, Dou Shen},pdfsubject={},
  pdfkeywords={Paper Reproduction, LLM Agents, Multi-Agent Systems, AI for Science}}

\newcommand{\sysname}{PACT\xspace}
\newcommand{\sysabbr}{PACT\xspace}
\newcommand{\syslong}{Paper Implementation with Artifact Consistency and Traceability}
\newcommand{\syslongbold}{\textbf{P}aper Implementation with \textbf{A}rtifact
  \textbf{C}onsistency and \textbf{T}raceability\xspace}
\newcommand{\coa}{Implementation Traceability Graph\xspace}
\newcommand{\coaabbr}{ITG\xspace}
\newcommand{\sgr}{Graph-Guided Review\xspace}
\newcommand{\sgrabbr}{GGR\xspace}
\newcommand{\ledger}{stage record\xspace}
\newcommand{\ledgers}{stage records\xspace}
\newcommand{\Ledger}{Stage record\xspace}

\newcommand{\pb}{PaperBench\xspace}
\newcommand{\pbcd}{PaperBench Code-Dev\xspace}
\newcommand{\pos}[1]{\textcolor{red!70!black}{\ensuremath{+#1}}}
\newcommand{\negv}[1]{\textcolor{green!45!black}{\ensuremath{-#1}}}
\newcommand{\id}[1]{\texttt{\small #1}}

\input{sections/00_meta}
\input{sections/0_abstract}

\begin{document}
\maketitle

\input{sections/1_introduction}
\input{sections/2_related_work}

\input{sections/3_method}
\input{sections/4_experiments}
\input{sections/5_conclusion}

\section*{Impact Statement}
This work aims to make automated reproduction of machine-learning papers more
faithful, traceable, and consistent. Reliable reproduction helps researchers verify
published claims and build on prior work, and it is a prerequisite for
autonomous research agents that extend existing methods. The same capability
could be used to produce convincing but unfaithful reimplementations; the
artifact graph we propose is designed to make such deviations visible by
linking each reported result to the code and run that produced it. Our
system respects benchmark blacklists and does not access authors' original
code.

\FloatBarrier
\bibliographystyle{unsrtnat}
\bibliography{references}

\clearpage
\beginappendix
\input{sections/A_appendix}

\end{document}

%% file: sections/00_meta.tex
\title{PACT: \syslong}
\shorttitle{PACT}

\author[1]{Zan~Gong}
\author[2]{Zengle~Ge}
\author[2]{Yui~Lo}
\author[1]{Linyue~Pan}
\author[1]{Bihui~Yin}
\author[3]{Guangxi~Wan}
\author[1]{Jinduo~Zhang}
\author[4]{Zhirong~Zhang}
\author[2]{Mo~Zhao}
\author[2]{Annan~Li}
\author[2]{Dou~Shen}

\affiliation[1]{Tsinghua University}
\affiliation[2]{Famou Agent Team, Baidu AI Cloud}
\makeatletter
\gappto\sa@affiliationlist{\\\sa@affiliationformat[3]{Beihang University}}
\makeatother
\affiliation[4]{Chang'an University}

\keywords{Paper Reproduction, LLM Agents, Multi-Agent Systems, AI for Science}

%% file: sections/0_abstract.tex
\abstract{
Agentic systems increasingly automate stages of AI research, from literature
review and idea generation to experiment design and manuscript writing.
A critical step in this workflow is paper implementation, which turns a
research design produced by earlier stages into a deployable, experimentally
grounded code repository.
However, this long-horizon step receives little immediate feedback:
experimental outcomes arrive late and often without reference values, allowing
silent drift from the paper's requirements to accumulate undetected.
We introduce PACT, a multi-agent framework for traceable and consistent
long-horizon paper implementation.
Specifically, an \coa links paper specifications, stage artifacts, code, runs,
and result claims into a unified graph, while \sgr reviews each stage's
artifacts and checks their consistency with adjacent stages to intercept defects where
they arise.
Experiments across open-weight backbones show that PACT generally outperforms
both published multi-agent systems and same-backbone single-agent baselines
while turning silent deviations into checkable, routable defects.
}

%% file: sections/1_introduction.tex
\section{Introduction}
\label{sec:intro}

\begin{figure}[!htbp]
  \centering
  \includegraphics[width=\textwidth]{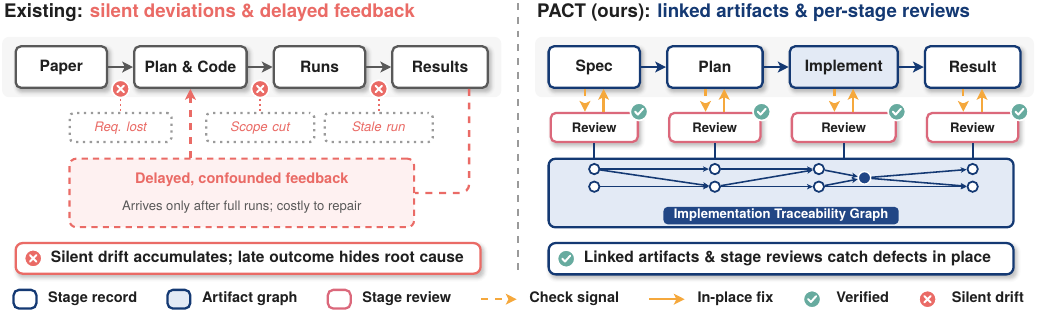}
  \caption{Long-horizon paper implementation without and with PACT. Left: in
  existing workflows, (i) deviations at stage hand-offs are silent, and (ii)
  feedback arrives only after full runs as a delayed, confounded signal. Right:
  PACT links specifications, plans, code, runs, and claims in an \coa and pairs
  each stage with \sgr to catch and repair defects in place.}
  \label{fig:challenges}
\end{figure}

Autonomous research systems now chain literature review, idea generation,
experimentation, and scientific writing into end-to-end pipelines
\citep{lu2026automation,rda2025,autoresearchclaw2026},
and self-improving agents rewrite their own code, harnesses, and skills, keeping
the changes that evaluate better
\citep{zhang2025darwin,zhang2026hyperagents,lee2026metaharness,karten2026continual,shen2026skilloptlite,zhang2026selfharness}.
Both depend on a step that is easy to overlook: implementation, which turns a
chosen design into running code and measured results. Once a method and an
experimental design have been selected, each later decision depends on whether
the code realizes that design and whether the reported findings come from
executing it. An incorrectly implemented idea may be discarded for the wrong reason, and a favorable result produced by a different method may provide false evidence for the original proposal.

% ---- [full definition -> Sec. problem (2b_problem.tex)] commented out 2026-11-03, kept for possible reuse ----
% We call this step \emph{paper implementation}: the faithful and auditable
% realization of a given research design, stated in a published paper or in an
% upstream draft. The deliverables are an executable repository, records of the
% experiments actually run, and a report grounded in their outputs. The method
% and the experimental commitments are fixed. Engineering choices may be refined,
% but replacing the scientific design is a decision for the upstream stage. This
% distinguishes paper implementation from discovery workflows, which may pivot to
% a new hypothesis after unsuccessful experiments \citep{autoresearchclaw2026}.
% It also separates fidelity from hypothesis support: a faithful implementation
% may yield a negative result, and favorable numbers do not excuse a
% methodological deviation.
% ---- end ----
We refer to this step as \emph{paper implementation}: realizing a given
research design, from a published paper or an upstream draft, as an executable
repository whose experiments are actually run and whose reported results come
from those runs, while the design itself remains fixed (Section~\ref{sec:method}).

Faithful implementation is long-horizon. Each requirement of the design has to
survive translation into a plan, several code modules, experiment
configurations and a result report, over many hours of work. Intuitively,
effective paper implementation should be traceable (\textit{i.e.}, each
reported result links back to the specification, code, and run that produced
it) and consistent (\textit{i.e.}, each stage's artifacts agree with upstream
commitments before downstream work builds on them). However, two properties of
this process make it difficult for current coding agents
(Figure~\ref{fig:challenges}).

\textbf{(i) Failures and deviations are silent.} A required baseline can drop
out of the executed schedule, an approximation can replace a specified
computation, or a reported result can outlive the code version that produced
it. None of these necessarily causes a crash: the code still runs and the pipeline
still produces numbers. PaperBench exposes such failures at repository and
experiment scale \citep{starace2025paperbench}, and ResearchCodeBench and
SciCoQA document them at the level of individual code snippets and
paper--code discrepancies \citep{hua2025researchcodebench,scicoqa2026}. Unit
tests and reduced-scale runs catch some of them but cannot establish fidelity
to the design. A draft often has no reference results to compare with, and
even when reference results exist, matching a published number does not show
that the intended method was implemented.

\textbf{(ii) Feedback is delayed.} AiScientist identifies delayed, confounded
feedback as a central challenge of long-horizon ML research engineering
\citep{aiscientistengineering2026}. The evidence that settles whether an
implementation is right, namely the full-scale experiments and their results,
arrives only after long runs. By then the plan and code that determine those
results have been written, and later work has been built on them. As a result,
a defect introduced early is expensive to repair. The late signal is also
confounded: a wrong number can come from a dropped requirement, a faulty module,
an undertrained run or a broken environment, and the outcome alone does not say
which. Some defects surface later still. Setup and download code that never
runs during development is exercised only when the submission is rerun in a
fresh environment, where a single broken path can invalidate all experiments.

Together, these properties mean that defects must be caught at the stage that
introduces them, turning delayed, confounded outcome feedback into immediate,
attributable cross-stage checks. Recent systems make parts of the path from
requirements to results explicit. AiScientist
stores analyses, plans and logs as persistent files shared among its agents
\citep{aiscientistengineering2026}, but these files are not tied to individual
requirements of the paper. ReproAgent derives contracts from the paper to guide
code generation \citep{reproagent2026}, and PaperCompiler compiles the paper
into repository-level specifications \citep{papercompiler2026}. Both concentrate on producing faithful code and do not propagate the
requirements forward into experiment execution and the reported results.
DeepCode's final verification stage checks the generated repository against its
blueprint and runs it in a sandbox \citep{li2025deepcode}. SigRef checks an
initial implementation against criteria drawn from the paper and iteratively
revises the code \citep{zhou2025repro}. Both check the finished outputs
at the end rather than the intermediate stages where a requirement is dropped
or a result goes stale. Therefore, a mechanism is needed to keep an
implementation consistent while it is being built.

We present \textbf{PACT} (\syslongbold), a multi-agent framework for
traceable and consistent long-horizon paper implementation, built around two
mechanisms.

For traceability, the
\textbf{\coa (\coaabbr)} connects each requirement of the paper to the plan
items, code, experiment runs and result claims that serve it, so that each
result can be traced back to the requirement it addresses. For artifact
consistency, the
\textbf{\sgr (\sgrabbr)} ends each stage with a review guided by this graph:
the reviewer checks that the stage's own artifacts are correct and that their
links to the records of adjacent stages hold, returning detected defects for repair before the next stage builds on those artifacts. Together, \coaabbr and \sgrabbr make long-horizon
paper implementation both traceable and consistent.

We evaluate PACT on published papers under the full PaperBench protocol, in
which the submitted repository is re-run in a fresh environment and an LLM
judge grades code, execution, and agreement with the published results
\citep{starace2025paperbench}. Our main contributions are as follows:
\begin{itemize}[leftmargin=*]
  \item We propose \textbf{\sysname}, a stage-structured multi-agent framework
  for long-horizon paper implementation that maintains cross-stage consistency
  from paper specifications to reported results while the system is being built.
  \item We couple the \textbf{\coa (\coaabbr)} with \textbf{\sgr (\sgrabbr)}:
  \coaabbr turns cross-stage dependencies into explicit graph links, and
  \sgrabbr reviews each stage using those links to catch and route silent
  deviations in place.
  \item Across the 20 \pb papers on two open-weight backbones, \sysname reaches
  43.93 with GLM-5 (+10.20 over published AiScientist on the same backbone and
  official judge) and 49.96 with DeepSeek-V4-Flash (+9.11 over same-backbone
  BasicAgent). Controlled component-removal studies and stage-review logs
  further show how \coaabbr and \sgrabbr complement each other to intercept
  silent deviations.
\end{itemize}

%% file: sections/2_related_work.tex
\section{Related Work}
\label{sec:related}

\textbf{End-to-end research agents.}
End-to-end research agents automate the research process as a whole. Systems
such as The AI Scientist, Agent Laboratory, AI-Researcher and InternAgent guide
a research idea through literature review, implementation and experiments, and
often into a written paper
\citep{lu2026automation,schmidgall2025agentlaboratory,tang2025airesearcher,internagent2025}.
Some add agentic tree search over experiments, memory that transfers lessons
across runs, or multi-agent debate with human-in-the-loop collaboration
\citep{yamada2025aiscientistv2,lyu2026evoscientist,autoresearchclaw2026}.
Others concentrate on experimentation, separating research from development
\citep{rda2025} or enforcing rigor at each experimental step
\citep{kon2025curie}, and ScientistOne makes each claim of its generated paper
traceable to supporting evidence \citep{scientistone2026}. In these systems,
implementation serves discovery: the code is judged by the outcomes it
produces, and a disappointing experiment can lead the system to refine or
replace its hypothesis \citep{autoresearchclaw2026}. Yet analyses of such
systems identify implementation and verification as their main points of
failure \citep{zhu2025aiscientistsfailstrong,whynotscientists2026}. \sysabbr
targets this step specifically: it keeps the design fixed and checks that each
stage remains consistent with it. Therefore, it can serve as the implementation
stage of such a pipeline.

\textbf{Paper-to-code agents.}
Paper-to-code agents address implementation directly. Workflow systems
decompose it into fixed stages such as planning, analysis and coding, guided by
repository blueprints, multimodal parsing of the paper or the lineage of its
citations \citep{autop2c2025,autoreproduce2025,seo2026paper2code}. Others
extract fine-grained requirements from the paper, either to verify and refine
the generated code \citep{zhou2025repro} or to constrain its generation
\citep{papercompiler2026}. Agentic systems give the agent more control, through
information-flow management, hierarchical supervision, state-aware replanning
or evolved harnesses \citep{li2025deepcode,hiras2026,deeprepro2026,milo2026},
and replication frameworks organize the work around the claims of the paper
\citep{paperreplication2026,veritas2026}. ReproAgent goes further by binding
the paper's requirements into file-level contracts and reviewing the generated
repository for compliance with them in a final repair stage \citep{reproagent2026}.
However, most of these systems are evaluates statically on \pbcd without executing
the delivered code from scratch in a clean environment.

Among systems targeting the full \pb execution protocol, Repo-To-Skill takes an
orthogonal knowledge-centric route, distilling related open-source repositories
offline into a reusable skill library for a fixed agent harness
\citep{repotoskill2026}. Closer to our architecture, CLEAR compiles the paper
into structured execution specifications to guide hierarchical planning and
localized repair \citep{clear2026}, and AiScientist coordinates specialist
agents through File-as-Bus, a workspace of durable artifacts that propagate state
across stages \citep{aiscientistengineering2026}. However, neither system maintains
end-to-end links from individual paper specifications through code to
versioned experiments and results or reviews adjacent artifacts at each stage
boundary. \sysabbr provides both: an \coa connecting each specification to the
code, runs and results that serve it, paired with \sgr to make long-horizon
paper implementation traceable and consistent before downstream work builds on
it.

%% file: sections/3_method.tex
\newpage
\section{\sysname: Consistency-Guided Long-Horizon Paper Implementation}
\label{sec:method}
% Previous version (record fields, coverage rules, modes in the main text):
% _old/3_method_before_formal_20261104.tex. Engineering details now live in App. A.

\begin{figure}[!t]
  \centering
  % trim (left bottom right top, in pt of the 960x540 slide) removes the slide's blank top and bottom bands
  \includegraphics[width=\textwidth,trim=8 80 3 90,clip]{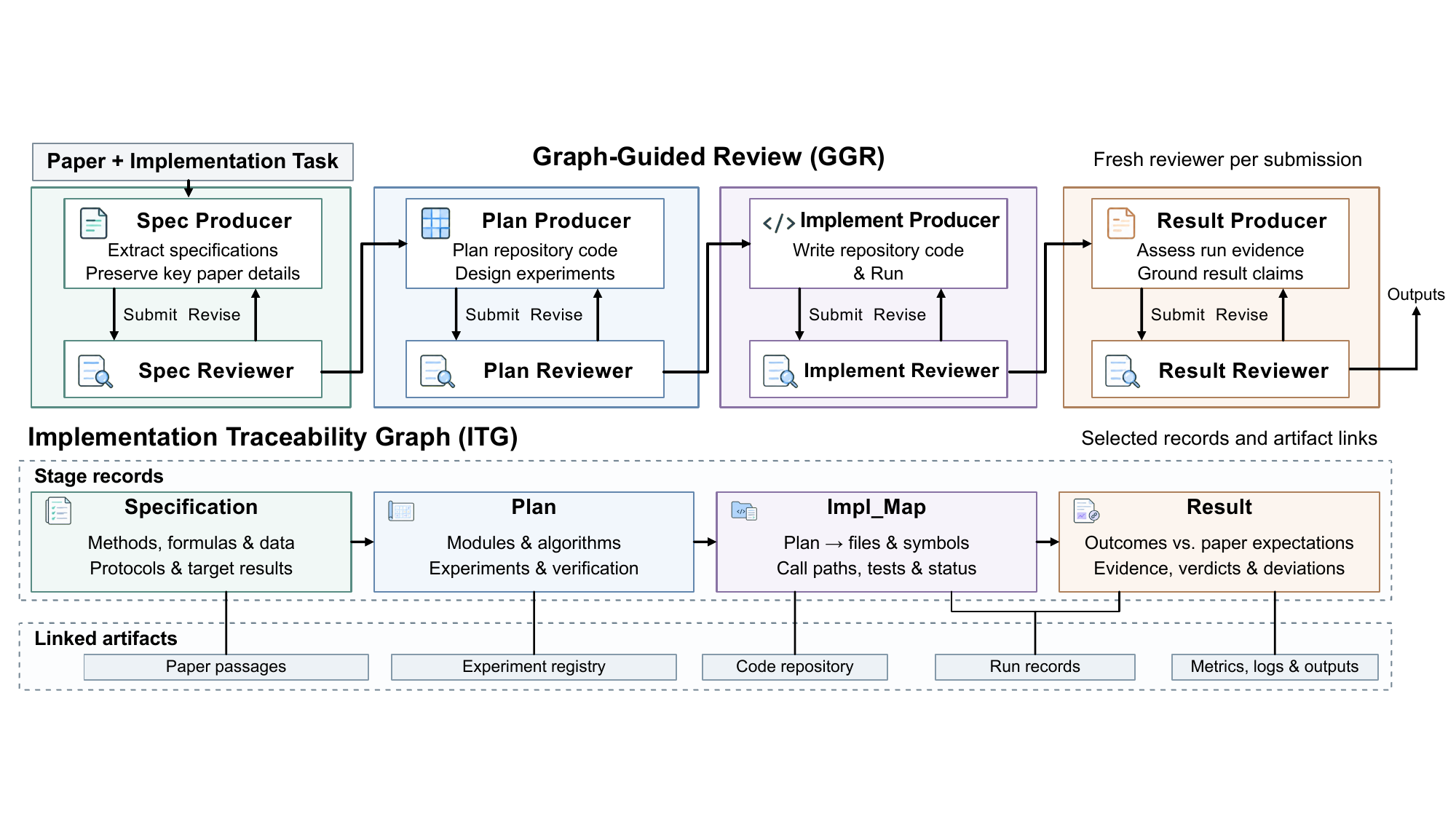}
  \caption{Overview of \sysabbr. Each stage pairs a producer with a separate
  reviewer session; local revisions address review findings before handoff.
  Review combines deterministic checks with assessment in a seperate session (\sgr). The
  lower panel illustrates selected stage records and linked artifacts in the
  \coa (\coaabbr), connecting specifications, plans, code, run evidence and
  result claims. The shared bracket denotes run evidence referenced by both
  implementation and result records. Cross-stage repair routes are omitted for
  clarity.}
  \label{fig:framework}
\end{figure}

\textbf{Task formulation.}
We define paper implementation as the process by which an
agent constructs and delivers a runnable system from a given research
design $D$, stated in a published paper or an upstream draft. The agent
may specify missing implementation details, but must preserve the
methodological and experimental requirements that $D$ specifies.

Given $D$, an execution contract $\mathcal{C}$, and a development budget
$B$, an implementation agent $\mathcal{A}$ produces a runnable system $S$:
\begin{equation}
  S = \mathcal{A}(D,\mathcal{C};B).
  \label{eq:method:task}
\end{equation}
The contract $\mathcal{C}$ specifies the target environment, permitted
resource access, execution limits, and delivery interface. The objective
is to realize $D$ within $B$ and make the correspondence
between the design and the delivered system inspectable. The delivered
system must support independent execution in a clean environment
satisfying $\mathcal{C}$, without further agent intervention.

\textbf{Overview.}
\sysabbr is a multi-agent framework for traceable and consistent paper
implementation, driven by a single controller. It divides
paper implementation into four stages (Spec, Plan, Implement, and Result;
(Figure~\ref{fig:framework}), each served by two agents. The producer of a
stage carries out its work and writes the artifacts it owes; its reviewer checks
those artifacts for accuracy and for consistency with the stages before it.
Among those artifacts is the stage's structured \ledger, and together the four
\ledgers connect the artifacts into the \coa (Table~\ref{tab:method:stages}).

The four stages follow the order in which the commitments of $D$ narrow
(Equation~\ref{eq:method:task}). \textbf{Spec} fixes what must hold, recording the
paper's methodological and experimental requirements with source excerpts from
the paper. \textbf{Plan} settles how each specification will be
met, pairing it with an approach, verification criterion, and experiment
protocol. \textbf{Implement} realizes the plans as code and runs reduced-scale
qualification experiments (Appendix~\ref{app:impl:qualify}), binding each output to
the code version that produced it. \textbf{Result} closes the loop by assessing
the available evidence relative to evaluation targets and disclosing open
deviations. No stage may rewrite upstream commitments, and each stage leaves
behind records that let the next be verified with respect to it.

\textbf{The implementation loop.}
The controller advances a run on typed events from the two roles. When a
producer submits its stage, the controller dispatches a reviewer to inspect the
submitted and upstream artifacts. If the reviewer accepts the submission, the stage advances and any open major findings are
carried forward as deviations for Result to disclose. However, if the reviewer rejects the submission, findings return to the submitting stage, or, when a
critical defect or accepted producer blocker originates upstream, trigger backtracking to
Plan or Implement within a bounded per-edge budget. Once frozen, Spec is never
reopened, preventing agents from altering the paper's original requirements and
ensuring that the final results correspond to the source paper rather than a
modified specification. If a stage exhausts its repair
budget, the controller advances and logs open findings as deviations.

\begin{table}[t]
  \centering
  \small
  \caption{Each stage writes a \ledger, and together they connect the
  stage artifacts into the \coa. \sgr checks these records at each
  submission.}
  \label{tab:method:stages}
  \setlength{\tabcolsep}{3pt}
  \renewcommand{\arraystretch}{1.12}
  \begin{tabularx}{\columnwidth}{@{}l>{\raggedright\arraybackslash}X@{}}
    \toprule
    Stage & \Ledger and the units it records \\
    \midrule
    Spec & Specifications and evaluation targets, anchored in $D$. \\
    Plan & Implementation and experiment plans, linked to the
    specifications they address. \\
    Implement & Code and execution configurations, linked to plans;
    observed runs and outputs, linked to the code versions they executed. \\
    Result & Assessments of the available evidence, linked to evaluation
    targets and to the runs and outputs they cite. \\
    \bottomrule
  \end{tabularx}
\end{table}

Two properties of this arrangement matter below. A producer may write only
the files of its own stage, and a stage's records remain frozen once they pass
review unless the controller reopens that stage. As a result, implementation
decisions are separated from the authority to revise the requirements they
serve, and a defect in a plan cannot be absorbed by a workaround in the code
that consumes it. The controller itself only orchestrates; the semantic
judgments come from reviewers that inspect the artifacts. Because both
mechanisms below read and write the same \ledgers, a single set of records
supports both the links that review follows and the repairs it triggers.
Appendix~\ref{app:impl} describes the controller, review rules, and record
formats.

\subsection{\coa}
\label{sec:method:coa}

\begin{figure}[!t]
  \centering
  \includegraphics[width=\textwidth]{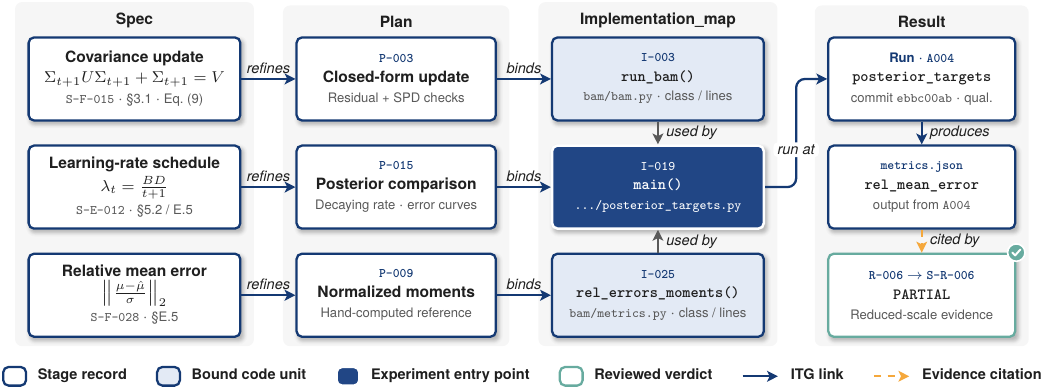}
  \caption{An illustrative \coaabbr subgraph for the BaM implementation task.
  Specifications are refined into plan entries with implementation approaches
  and verification criteria; implementation bindings connect plans to code
  locations. The \texttt{main()} function in \texttt{posterior\_targets.py}
  uses the covariance update and computes the relative mean error.
  \texttt{PARTIAL} indicates that the result from the reduced-scale run
  partially supports the paper's claim.}
  \label{fig:bam-itg-example}
\end{figure}

We represent the correspondences maintained during implementation as a
directed graph $G=(V,E)$, where $V$ is the set of inspectable artifact units
(specifications, plan entries, experiment definitions, code or configuration
units, recorded runs, outputs, and assessments) and $E$ is the set of directed
relationships between them. Nodes are neither agents nor whole stages, and a
single file can contain many of them. Specifically, each node records an
identifier, the content or location to be inspected, and its stage, type, and
version context, whereas an edge simply captures a logical reference between
two units. These correspondences form a graph rather than a chain because dependencies branch and artifact unit can be shared: one specification may require several implementation units,
one code unit may serve several specifications, and experiments may depend on
one another.

The construction below follows from three properties we require of an
implementation and its record. \textbf{Coverage:} each requirement the paper
implies remains reachable from a plan entry and from an experiment, so that a
required comparison cannot be dropped without leaving a trace.
\textbf{Realization:} the paper's own wording travels with each specification,
so that a change of meaning in the implementation appears as a textual
difference rather than an unrecorded choice. \textbf{Traceability:} each
reported result is bound to the run, command, and commit that produced it, so
that an output cannot be cited without the process behind it.
Section~\ref{sec:method:sgr} turns each of them into a review target.

\textbf{Construction through \ledgers.}
The producers build $G$ stage by stage. Spec adds specifications anchored to
locations in $D$, each with the precision it must respect. Plan adds plan
entries and experiment definitions that cite the specifications they
address, and keeps each specification's text together with the proposed
approach, so that a change of meaning shows up as a textual difference.
Implement adds implementation bindings: edges from plan entries to the code
units that realize them. It also adds each experiment's execution configuration and observed runs, each tied to a code version and its outputs. Result adds assessments that cite the evaluation targets and the
runs and outputs they rely on. Repairs may revise units and links. Therefore,
the graph need not grow monotonically. $G$ is the logical graph defined by
these identities and references across the \ledgers. The \ledgers themselves
are structured text files, which agents edit directly, reviewers read, version
control compares, and the controller indexes deterministically
(Appendix~\ref{app:impl:records}).

\textbf{An example.}
Figure~\ref{fig:bam-itg-example} shows part of the graph built for BaM
\citep{cai2024bam}. Read from left to right, it follows one thread of the
implementation. Three specifications enter from the paper: the covariance
update (S-F-015), the learning-rate schedule (S-E-012) and the relative error
used for evaluation (S-F-028). Each is refined by a plan entry that states an
approach and how it will be verified, and each plan entry is bound to the code
unit that realizes it. The experiment entry point bound to the schedule plan calls
both the update and the metric function, making a change to either visible from
the experiment. This entry point is linked to a qualification run at a recorded
commit, the run to the metrics file it produced, and the file to the result
assessment that cites it. The assessment is \texttt{PARTIAL}: the
reduced-scale run reproduces only part of the target claim, and the record
states this scope explicitly for review.

\subsection{\sgr}
\label{sec:method:sgr}
Once a stage's producer submits its work, \sgr is initiated to inspect the
submission relative to its requirements and return a verdict and findings
(Equation~\ref{eq:method:review}). Its semantic review runs in a fresh agent
session without the producer's conversation history. Optimized language models
have been
shown to defend flawed outputs with selectively chosen evidence and to
satisfy stated checks without meeting their intent
\citep{wen2025mislead,baker2025monitoring}, and verification failures are a
recurring weakness in multi-agent pipelines \citep{cemri2025multiagentfail}.
Therefore, the reviewer checks the submission relative to the stage records,
which the producer did not write for the occasion.

At the $k$-th submission of stage $s$, a review is the mapping:
\begin{equation}
  (d_k,F_k)=\operatorname{Review}\bigl(G_k^{(s)};\mathcal{C}\bigr),
  \label{eq:method:review}
\end{equation}
which takes the graph view $G_k^{(s)}$ of the current \coa and the execution
contract $\mathcal{C}$, and returns verdict $d_k\in\{\textsc{Pass},
\textsc{Pass-with-issues}, \textsc{Reject}\}$ and finding set $F_k$. Each
finding $f\in F_k$ names a concrete location, the observed discrepancy, a
severity (\texttt{critical}, \texttt{major}, or \texttt{minor}), a required
fix, and the owner stage whose \ledger must change. An open \texttt{critical}
finding or accepted blocker forces \textsc{Reject}, returning repair to the
submitting stage or an allowed upstream owner stage;
\textsc{Pass-with-issues} allows advancement with open \texttt{major} findings
logged as deviations for Result to disclose (Appendix~\ref{app:impl:review}).
Later reviews check how earlier findings were addressed.

The review process proceeds in two steps. A deterministic gate first checks that the
stage records exist and are well formed, screening out formatting problems and
missing entries before any agent session is spent on them; a gate failure
returns concrete diagnostics to the producer. The reviewer then addresses the
semantic questions that the gate cannot resolve by opening the source artifacts named
by the graph view. For an Implement submission, these are the code at the
submitted revision and the execution configuration, which the reviewer may run
as tests or reduced-scale commands in a scratch environment. The reviewer follows two
principles. \textbf{Internal consistency} requires the stage record to be
internally coherent and to agree with the other artifacts the stage produced.
\textbf{Adjacent-stage consistency} requires the record to cover each required
unit of the upstream stage, with each entry satisfying the upstream requirement
it addresses.

%% file: sections/4_experiments.tex
\section{Experiments}
\label{sec:exp}

Our experiments evaluate \sysabbr along three dimensions.
First, we measure end-to-end implementation quality compared with published
systems and a same-backbone single-agent baseline.
Second, we examine the complementary roles of \coaabbr and \sgrabbr through the
failure modes that surface when either mechanism is removed.
Third, we analyze the defects intercepted by \sgrabbr across the benchmark and
the review principles responsible for catching them.

\subsection{Setup}
\label{sec:exp:setup}

\textbf{Evaluation task and benchmark.}
Evaluating the implementation stage of a research pipeline requires testing
whether an agent can faithfully realize a specified design, execute it
end-to-end, and produce verifiable results rather than only generating
isolated code or open-ended research outputs. We adopt \pb
\citep{starace2025paperbench}, a paper reproduction benchmark whose protocol
closely aligns with this setting: the agent receives a paper and its
authors' addendum, must implement the work from scratch, deploy it, and run
its experiments without accessing the authors' code. The published results
serve as the reference, and rubrics assess whether the agent's submission
reproduces them. The benchmark covers 20 ICML~2024 papers with 8{,}316
gradable requirements. Each requirement is a leaf of one of three types:
Code Development (CD), which grades code structure; Code Execution
(CE), which checks whether experiments run; or Result Analysis (RA), which
measures agreement with the paper's numbers. In the full \pb pipeline, the judge grades each leaf.
The replication score is the recursively weighted average of the leaf scores.

\textbf{Baselines, backbones and grading protocols.}
We evaluate \sysabbr in two head-to-head comparisons using one NVIDIA
A800-80GB GPU, 8 CPU cores, a 24-hour development budget (plus a 24-hour
clean-environment reproduction budget), and GPT-5.4 \citep{openai2026gpt54} as
the judge. First, to compare with published systems, we run
\sysabbr with GLM-5 \citep{zeng2026glm} for all producer and reviewer roles.
We compare these runs with BasicAgent, IterativeAgent \citep{starace2025paperbench},
and AiScientist \citep{aiscientistengineering2026} on GLM-5, as well as Codex
with GPT-5.5 (xhigh) as a frontier-harness reference reported by
\citet{aiscientistengineering2026} under the same budgets and judge
(Table~\ref{tab:main}).
Second, to test backbone generality and isolate the framework from the model,
we run both \sysabbr and
the official BasicAgent solver with DeepSeek-V4-Flash
\citep{deepseekai2026v4release} under identical hardware and budgets
(Table~\ref{tab:basicagent}).

\newpage
\input{tables/main_comparison}

\subsection{Main Results}
\label{sec:exp:main}

\textbf{Comparison with published GLM-5 systems.}
Table~\ref{tab:main} compares \sysabbr with the published baselines from
\citet{aiscientistengineering2026} under the same judge and budget. With
GLM-5, \sysabbr achieves 43.93, outperforming AiScientist by 10.20 points
(leading on 16 of 20 papers) and BasicAgent by 21.35 points. The mean scores
for \sysabbr (GLM-5) and the Codex/GPT-5.5 frontier-harness reference are
43.93 and 29.45, respectively. These gains extend across the benchmark rather than being driven by isolated outliers, spanning diverse domains and multi-experiment protocols.

\input{tables/basicagent_dsv4f}

\textbf{Same-backbone comparison with DeepSeek-V4-Flash.}
Table~\ref{tab:basicagent} isolates the contribution of the framework under the
same DeepSeek-V4-Flash backbone, hardware, and grading protocol.
\sysabbr reaches 49.96 compared with 40.85 for BasicAgent, gaining 9.11 points
and leading on 15 of 20 papers at an average rollout cost of \$4.3 per paper
(Appendix~\ref{app:eval}).
Two recurring patterns explain where \sysabbr improves over BasicAgent.
First, on complex, multi-component papers requiring many baselines,
environments, or experiment grids (e.g., \texttt{all-in-one}, \texttt{sapg},
\texttt{sequential-neural}, \texttt{bbox}, and \texttt{rice}), \sysabbr
helps avoid scope loss and missing deliverables: under a
single-conversation workflow, BasicAgent often drops secondary scope under time
pressure or fails to deliver a runnable entry point, whereas \coaabbr binds
paper obligations end-to-end into the experiment registry and generated
\texttt{reproduce.sh}.
Second, even where BasicAgent implements the core pipeline (e.g.,
\texttt{adaptive-pruning}, \texttt{bridging-data-gaps}, and \texttt{fre}),
\coaabbr preserves exact formulas, hyperparameters, and task definitions across
stages---using explicit adjudications when external dependencies are
unavailable---while \sgrabbr intercepts unwired modules and formula drift before
delivery.

Conversely, on the five papers where \sysabbr scores lower than
BasicAgent---primarily \texttt{mechanistic-understanding} and
\texttt{stay-on-topic}, with smaller gaps on \texttt{pinn}, \texttt{lbcs}, and
near-parity on \texttt{stochastic-interpolants}---the deficits arise from
execution-stage engineering and runtime issues during clean-environment
reproduction (such as container interruptions, filesystem constraints,
environment-setup failures, and resource-contention errors) that prevented the
delivered code from completing within the 24\,h budget
(Appendix~\ref{app:loss}).

\subsection{Complementary Roles of \coaabbr and \sgrabbr}
\label{sec:exp:case}

\begin{figure}[!t]
  \centering
  \includegraphics[width=\textwidth]{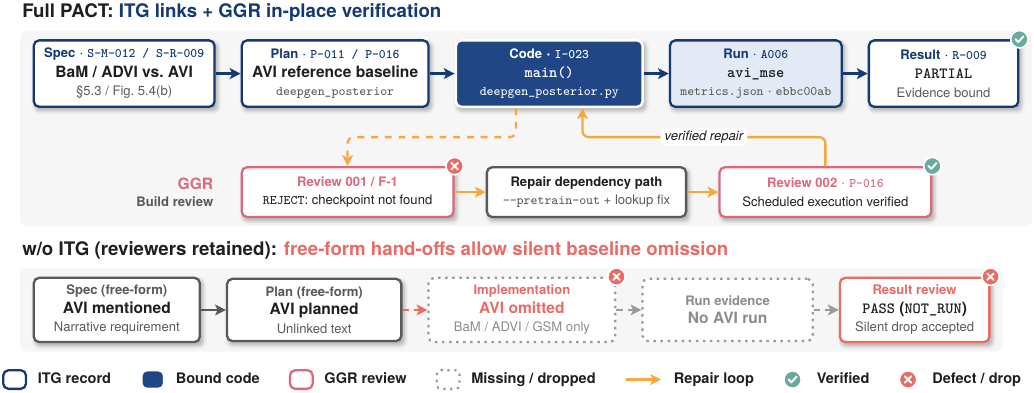}
  \caption{Trace of the amortized variational inference (AVI) baseline
  comparison in the \pb \texttt{bam} task \citep{cai2024bam}. Top and middle:
  in the full run, \coaabbr links the required comparison
  (\texttt{S-M-012}/\texttt{S-R-009} and \texttt{P-011}/\texttt{P-016}) to its
  registered experiment (\texttt{deepgen\_posterior}), implementation binding
  (\texttt{I-023}), versioned qualification run (\texttt{A006}), and result
  record (\texttt{R-009}, \textsc{Partial}); \sgrabbr build review~001 catches
  a checkpoint-path failure and review~002 verifies the repair. Bottom: without
  \coaabbr (reviewers retained, cross-stage repair removed), AVI is planned in
  free-form text, omitted from code (leaving only BaM, ADVI, and GSM), and
  disclosed as \texttt{NOT\_RUN} under an accepted Result review (\textsc{Pass}).}
  \label{fig:bam-avi-case}
\end{figure}

\coaabbr and \sgrabbr play complementary roles in \sysabbr: \coaabbr records
structured obligations, verification criteria, and cross-stage repair paths,
while \sgrabbr reviews whether downstream code, scripts, and runs actually honor
them. Therefore, removing either mechanism exposes what the remaining one alone
cannot prevent: without \coaabbr, reviewers face unstructured documents with no
formal obligation to enforce and no repair route to upstream stages; without
\sgrabbr, \coaabbr records commitments without verifying that they are
realized. To examine these failure modes, we conduct controlled
component-removal trials on \texttt{bam} \citep{cai2024bam}, a representative
\pb task (Table~\ref{tab:main}) on black-box variational inference. This task
specifies exact protocol parameters and requires comparing BaM with multiple
baselines, including ADVI, GSM, and amortized variational inference (AVI). We
compare the full system with variants that remove \coaabbr or remove \sgrabbr
on Spec, Plan, and Implement.

\textbf{Removing \coaabbr (reviewers retained).}
When stage records, the merged experiment registry, generated schedules, and
cross-stage backtracking are replaced by free-form documents, all three
properties of Section~\ref{sec:method:coa} fail visibly even though each stage is
still reviewed.
\textbf{Coverage:} Figure~\ref{fig:bam-avi-case} traces the required
comparison of BaM and ADVI with the AVI baseline in the deep generative
posterior experiment (\S5.3).
In the full run, \coaabbr connects this specification through its plan entries
and registered experiment to the experiment entry point, a reduced-scale qualification run
that records the AVI reconstruction-error metric (\texttt{avi\_mse}), and a
\textsc{Partial} result record.
During build review, \sgrabbr catches a broken checkpoint path and verifies its
repair before the stage advances.
In contrast, without \coaabbr, the AVI baseline is mentioned in the free-form
specification and plan yet silently omitted from the code, leaving only BaM,
ADVI, and GSM.
With no run evidence, Result reports AVI as \texttt{NOT\_RUN} and still
receives a \textsc{Pass}, because no explicit link requires planned baselines
to survive into implementation.
\textbf{Realization:} the paper specifies a decaying learning-rate schedule
$\lambda_t{=}BD/(t{+}1)$, which full \sysabbr implements exactly as specified.
Without \coaabbr, the implementer replaces this schedule with a constant
multiplier selected from a pilot grid.
Although the reviewer notices the change, the requirement appears only in plain
text without a formal exactness constraint from Spec; it is therefore logged as
a non-blocking major finding and the stage advances.
\textbf{Traceability:} the same run hand-writes \texttt{reproduce.sh} so that a
short quick-mode sanity check writes directly into the main output directory.
As a result, subsequent full-scale experiments mistake those sanity-check files
for completed outputs and skip full execution.
With \coaabbr, each of these drifts surfaces as a broken graph link rather than
an undocumented omission.

\textbf{Removing \sgrabbr (\coaabbr retained).}
In this variant, we retain \coaabbr and the final Result review and remove the
Spec, Plan, and Implement reviews. The resulting submissions illustrate two
representative failure modes.
\textbf{Clean-environment delivery failures:} reproducing \texttt{bam} requires
preparing external benchmarks (such as CIFAR-10 and PosteriorDB) from scratch
in a fresh container. Without \sgrabbr, delivery scripts and execution configs
repeatedly fail during setup or execution because of mismatched data paths, broken download URLs, relative build paths, or overly tight timeout limits. Local rollout caches mask these defects until clean-environment evaluation.
\textbf{Silent realization failures:} even when environment setup succeeds,
unreviewed plans and code allow degenerate models to pass undetected.
In \texttt{bam}'s deep generative experiment, the pipeline first pretrains a
CIFAR-10 variational autoencoder (VAE) and then runs latent posterior
inference.
Without \sgrabbr, the unreviewed plan checks only output tensor shapes rather
than training convergence.
At the same time, an unchecked bug in the VAE (unbounded decoder variance paired with a scale mismatch between the negative log-likelihood (NLL) and KL-divergence loss terms) causes the decoder to collapse to a near-constant
image, degrading posterior reconstruction MSE from 0.019 in full \sysabbr to
0.157.
Because the script exits without error and satisfies each shape check, the run
is marked \texttt{completed} and delivered unchanged.

Taken together, these two ablations illustrate how \coaabbr and \sgrabbr
complement each other: \coaabbr supplies structured obligations, verification
criteria, and repair paths that make reviews enforceable, while \sgrabbr
checks the delivered code, scripts, and run evidence relative to those commitments.

\subsection{What Defects Does \sgr Catch?}
\label{sec:exp:sgr}

Inspecting the review logs across both backbones shows that \sgrabbr catches two
distinct classes of silent defects that normal code execution misses.

First, \textbf{cross-stage mismatches} occur when a stage's output runs cleanly
but quietly departs from upstream commitments in Spec or Plan.
By comparing submitted code and results with upstream stage records, \sgrabbr
repeatedly catches formulas or hyperparameters that drift from the paper,
required baselines dropped from the execution schedule, and broken environment
or entry-point scripts.
For example, in \texttt{adaptive-pruning}, the training code computed pruning
masks on each forward pass but never applied them to the model weights; the
script ran without error, yet \sgrabbr rejected the implementation by verifying
the code path relative to the planned pruning specification.
Similarly, in \texttt{bridging-data-gaps}, the plan required four ablation
variants, but the launch script executed only one; \sgrabbr caught the missing
modes by checking the script relative to the experiment registry.

Second, \textbf{within-stage inconsistencies} occur when artifacts produced
inside the same stage contradict one another, such as hard-coded placeholder
metrics, dead code never called by \texttt{reproduce.sh}, or claims that
disagree with the stage's own run logs.
For instance, in \texttt{sequential-neural}, the result report claimed a budget of $10{,}000$ simulations to match the paper's table, whereas the linked run log
recorded \texttt{budget=1000}; \sgrabbr rejected the report by cross-checking
the claimed number with the linked run artifact.
In each case, \sgrabbr turns a silent deviation into a concrete repair before the next stage builds on the affected artifact (see Appendix~\ref{app:impl:review} for review and
backtracking statistics).

%% file: tables/main_comparison.tex
\begin{table}[!t]
\centering
\small
\begin{threeparttable}
\caption{Overall replication score (\%) on the full \pb protocol (GPT-5.4 judge,
24\,h). The second header row gives each system's
rollout model. All baseline columns are as reported by
\citet{aiscientistengineering2026} under the same judge and budget; Codex is a
frontier-harness reference. Bold marks the highest score in each row.}
\label{tab:main}
\setlength{\tabcolsep}{5pt}
\begin{tabular*}{\textwidth}{@{\extracolsep{\fill}}l ccccc@{}}
\toprule
Paper & BasicAgent & IterAgent & AiScientist & Codex & \textbf{\sysabbr}\tnote{a} \\
      & (GLM-5) & (GLM-5) & (GLM-5) & (GPT-5.5) & (GLM-5) \\
\midrule
adaptive-pruning           & 30.82 & 11.93 & 33.26 & 33.42 & \textbf{54.40} \\
all-in-one                 & 33.78 & 44.43 & 49.47 & 52.93 & \textbf{61.83} \\
bam                        & 51.45 & 47.91 & 61.11 & 56.65 & \textbf{61.58} \\
bbox                       & 23.55 & 19.28 & 30.02 & 17.30 & \textbf{44.61} \\
bridging-data-gaps         & 9.80  & 12.50 & \textbf{26.46} & 14.16 & 23.51 \\
fre                        & 21.60 & 16.67 & 28.98 & 14.06 & \textbf{39.76} \\
ftrl                       & 3.71  & 6.70  & 8.34  & 1.50  & \textbf{19.21} \\
lbcs                       & 20.68 & 22.74 & 30.10 & 26.60 & \textbf{31.23} \\
lca-on-the-line            & 22.55 & 26.15 & 28.53 & 22.08 & \textbf{45.39} \\
mechanistic-understanding  & 32.49 & 34.96 & 40.55 & 45.67 & \textbf{65.58} \\
pinn                       & 22.18 & 25.77 & \textbf{58.76} & 40.64 & 53.02 \\
rice                       & 6.56  & 0.27  & 10.18 & 7.94  & \textbf{34.73} \\
robust-clip                & 22.43 & 27.56 & 28.66 & 30.14 & \textbf{52.68} \\
sample-specific-masks      & 36.93 & 41.26 & 44.13 & \textbf{57.11} & 53.79 \\
sapg                       & 6.99  & 4.95  & 31.69 & 18.15 & \textbf{32.44} \\
sequential-neural          & 27.20 & 35.53 & 49.32 & 41.67 & \textbf{68.34} \\
stay-on-topic              & 3.69  & 8.81  & 14.81 & \textbf{32.31} & 16.09 \\
stochastic-interpolants    & 32.18 & 28.06 & \textbf{42.10} & 41.12 & 35.78 \\
test-time-model-adaptation & 17.81 & 21.19 & \textbf{27.33} & 26.28 & 25.39 \\
what-will-my-model-forget  & 25.14 & 10.75 & 30.82 & 9.35  & \textbf{59.33} \\
\midrule
\textbf{Average}           & 22.58 & 22.37 & 33.73 & 29.45 & \textbf{43.93} \\
\bottomrule
\end{tabular*}
\begin{tablenotes}[flushleft]\footnotesize
\item[a] We scored the \sysabbr submissions using the official \pb grading logic for the full rubric.
\end{tablenotes}
\end{threeparttable}
\end{table}

%% file: tables/basicagent_dsv4f.tex
\begin{table}[!t]
\centering
\small
\begin{threeparttable}
\caption{Same-backbone replication score (\%) with DeepSeek-V4-Flash on the
full \pb protocol (GPT-5.4 judge).\tnote{b} $\Delta$ is the \sysabbr score minus the
BasicAgent score, in percentage points, using the values shown.}
\label{tab:basicagent}
\setlength{\tabcolsep}{4pt}
\begin{tabular*}{\columnwidth}{@{\extracolsep{\fill}}lccr@{}}
\toprule
Paper & BasicAgent & \textbf{\sysabbr} & $\Delta$ (pp) \\
\midrule
adaptive-pruning                     & 39.90 & \textbf{59.88} & \pos{19.98} \\
all-in-one                           & 25.68 & \textbf{63.21} & \pos{37.53} \\
bam                                  & 69.92 & \textbf{86.22} & \pos{16.30} \\
bbox                                 & 29.08 & \textbf{43.09} & \pos{14.01} \\
bridging-data-gaps                   & 27.81 & \textbf{37.79} & \pos{9.98} \\
fre                                  & 39.71 & \textbf{45.35} & \pos{5.64} \\
ftrl                                 & 14.50 & \textbf{19.07} & \pos{4.57} \\
lbcs                                 & \textbf{50.86} & 47.77 & \negv{3.09} \\
lca-on-the-line            & 26.61 & \textbf{41.20} & \pos{14.59} \\
mechanistic-understanding            & \textbf{77.97} & 52.79 & \negv{25.18} \\
pinn                                 & \textbf{46.08} & 41.77 & \negv{4.31} \\
rice                                 & 27.49 & \textbf{36.41} & \pos{8.92} \\
robust-clip                & 38.96 & \textbf{56.32} & \pos{17.36} \\
sample-specific-masks                & 60.87 & \textbf{67.94} & \pos{7.07} \\
sapg                                 & 13.85 & \textbf{51.28} & \pos{37.43} \\
sequential-neural                    & 52.95 & \textbf{78.79} & \pos{25.84} \\
stay-on-topic                        & \textbf{35.09} & 21.27 & \negv{13.82} \\
stochastic-interpolants              & \textbf{49.74} & 49.36 & \negv{0.38} \\
test-time-model-adaptation & 28.85 & \textbf{32.51} & \pos{3.66} \\
what-will-my-model-forget            & 61.01 & \textbf{67.23} & \pos{6.22} \\
\midrule
\textbf{Average (20 papers)}         & 40.85 & \textbf{49.96} & \pos{9.11} \\
\bottomrule
\end{tabular*}
\begin{tablenotes}[flushleft]\footnotesize
\item[b] In our GPT-5.4 evaluations, the official grader sometimes omitted
selected files when the file ranker returned paths with a leading slash.
For both systems, we strip that slash so the grader can read the files
relative to the submission directory.
\end{tablenotes}
\end{threeparttable}
\end{table}

%% file: sections/5_conclusion.tex
\section{Conclusion}
\label{sec:conclusion}

We presented PACT, a multi-agent framework for traceable and consistent
long-horizon paper implementation.
Specifically, PACT coupled the \coa (\coaabbr) with \sgr (\sgrabbr): \coaabbr
linked paper specifications, plans, code, runs, and result claims into a
unified graph, while \sgrabbr reviewed each stage boundary to intercept and
route defects where they arose.
Experiments across open-weight backbones showed that PACT generally
outperformed both published multi-agent systems and same-backbone single-agent
baselines. Controlled component-removal studies and multi-run review logs confirmed that \coaabbr and \sgrabbr complement each other to turn
silent failures into checkable, routable defects.\label{main-text-end}

%% file: sections/A_appendix.tex
\raggedbottom
\section{Implementation Details}
\label{app:impl}

This appendix summarizes the system configuration used for all reported runs:
controller orchestration and isolation (Appendix~\ref{app:impl:controller}), external
retrieval and blacklist enforcement (Appendix~\ref{app:impl:search}), stage record
schemas (Appendix~\ref{app:impl:records}), the Implement stage workflow
(Appendix~\ref{app:impl:implement}), and review and backtracking rules
(Appendix~\ref{app:impl:review}). Appendix~\ref{app:role-prompts} provides the
instructions for each producer and its reviewer.

\FloatBarrier
\subsection{Controller, Scope Isolation and Budgeting}
\label{app:impl:controller}

\textbf{Harness and models.}
All agent roles run on the \texttt{pi} coding-agent harness through an interface that allows other backends to be used. Within a run, one
backbone serves all producer and reviewer roles: GLM-5 or DeepSeek-V4-Flash
via an OpenAI-compatible endpoint. Each stage producer keeps a single
long-lived session, whereas each reviewer runs in a fresh session per
submission without the producer's conversation history.

\textbf{Event-driven orchestration.}
A deterministic controller that makes no LLM calls drives the pipeline from an
append-only event log, allowing run state to be replayed deterministically. Agents
never invoke one another directly; instead, role-scoped control actions
(\texttt{submit}, \texttt{checkpoint}, \texttt{review}) record an intent that
the controller validates and applies only after the agent's turn ends,
ensuring that a stage never advances mid-edit. Review receipts are validated before
recorded (required coverage rows, verdict--finding consistency, valid
severities, owner stages, locations, and existing run IDs); malformed receipts
are returned to the reviewer with diagnostics.

\textbf{Scope isolation.}
Write boundaries are enforced at both the tool-guard and OS-permission levels
rather than by prompting alone. Each producer may modify only its own stage
artifacts while active; once a stage passes review, its records are frozen
read-only unless explicitly reopened by the controller. Reviewers own no
workspace files: they inspect a read-only snapshot of the submitted commit and
report outcomes solely through structured review receipts. While an experiment
executes, edits to the code tree are blocked so that each run record binds
cleanly to the exact code revision executed.

\textbf{Time budgeting and bounded retries.}
All stages share a single 24-hour development clock. Each stage and Implement
mode receives a soft budget cap proportional to the total window, reserving
tail time for final delivery and Result synthesis. At high budget utilization,
the controller nudges active sessions with the remaining time so that producers
checkpoint and reviewers finalize receipts. Consecutive review rejections in
a stage or mode are capped at three rounds, after which the controller
advances while preserving the actual verdict and logging open findings as
deviations to avoid deadlock.

\FloatBarrier
\subsection{External Retrieval and Blacklist Enforcement}
\label{app:impl:search}

\textbf{Structured retrieval in Plan.}
While agents may use standard shell commands to install packages or fetch
datasets and pretrained models, structured external retrieval
(\texttt{web\_search}, \texttt{web\_fetch}, and \texttt{repo\_fetch}) is
integrated into the Plan stage so the planner can clarify underspecified
methodological details or adopt permitted public libraries during design.
Retrieved references are stored under \texttt{references/REF-NNN/} with a
summary, optional snippets or shallow clones, and controller-recorded
provenance (URL, commit, and fork check), then frozen read-only before
Implement begins. Code reused from a reference includes a
\texttt{ref: REF-NNN path:L$x$--L$y$} citation comment.

\textbf{Blacklist enforcement.}
To comply with the \pb prohibition on accessing authors' original code,
normalized repository and URL identities parsed from \texttt{blacklist.txt}
are enforced across all roles via four layers: (1) explicit role instructions
forbidding fetching, cloning, installing, citing, or reconstructing
blacklisted sources (including forks and mirrors); (2) tool and command guards
blocking arguments that reference blacklisted identities; (3) search-result
filtering, post-redirect URL checks, and GitHub fork-parent checks on
\texttt{web\_search}/\texttt{web\_fetch}/\texttt{repo\_fetch}; and (4) a
pre-delivery provenance audit over the submission and reference directory.

\FloatBarrier
\subsection{Stage Records}
\label{app:impl:records}

Table~\ref{tab:records} pairs each stage in Table~\ref{tab:method:stages} with
its \ledger in Figure~\ref{fig:framework}, listing the underlying files, entry
fields, and checks alongside the controller-maintained linked artifacts they
reference.
The controller parses \texttt{spec.md}, \texttt{plan.md}, and
\texttt{result.md} into structured indexes so cross-stage links can be checked
automatically.
During the Implement stage, the producer maintains
\texttt{implementation\_map.md} (\texttt{Impl\_Map}), linking each plan entry
to its target \texttt{path::symbol}, call path from the entry point,
verification test, and run bindings; reviewers use this map as a checklist to
inspect the actual code rather than relying on automated test-coverage metrics.
For experiments, the Plan producer defines the scientific protocol
(\texttt{experiment-plan.yaml}) and the Implement producer supplies the
execution configuration (\texttt{experiment-exec.yaml}); the controller merges
both files by ID into the \emph{Experiment registry} and generated schedule
(Appendix~\ref{app:impl:registry}), while an independent supervisor writes
version-bound \emph{Run records} referenced by both
\texttt{implementation\_map.md} and \texttt{result.md}.

\begin{table}[H]
\centering
\small
\caption{Schemas of the four \ledgers in the \coa and the
controller-maintained linked artifacts they reference
(Figure~\ref{fig:framework}). The deterministic gate (G) checks link presence
and structure; the reviewer (R) audits content and semantic consistency.}
\label{tab:records}
\setlength{\tabcolsep}{5pt}
\begin{tabularx}{\columnwidth}{@{}ll>{\raggedright\arraybackslash}Xl@{}}
\toprule
Stage (record) & Component & Entry fields & Check \\
\midrule
Spec (Specification) & \texttt{spec.md} & \id{S-*} id, type, exactness (\texttt{exact}/\texttt{minimum}/\texttt{range}/\texttt{qualitative}), source excerpt, paper anchor, required & G, R \\
\addlinespace[2pt]
Plan (Plan) & \texttt{plan.md} & \id{P-*} id, covered \id{S}-ids, source excerpt, approach, verification & G, R \\
 & \texttt{experiment-plan.yaml} & experiment id, \id{S}/\id{P}-ids, priority, seeds, protocol, expected outputs & G, R \\
\addlinespace[2pt]
Implement (Impl\_Map) & \texttt{implementation\_map.md} & \id{I-*} id, \id{P}-ids, \texttt{path::symbol}, call path, test, status; run bindings and \textsc{rerun}/\textsc{reuse} actions & G, R \\
 & \texttt{experiment-exec.yaml} & experiment id, full/qual commands, \texttt{code\_paths}, \texttt{depends\_on}, resource estimates & G, R \\
\addlinespace[2pt]
Result (Result) & \texttt{result.md} & \id{R-*} id, \id{S-R}-ids, run id, output path, value, verdict, deviations & G, R \\
\midrule
\multicolumn{4}{@{}l}{\textit{Controller-maintained linked artifacts (referenced by the \ledgers above)}} \\
\addlinespace[2pt]
Experiment registry & \texttt{.impl/registry.json} / schedule & merged \texttt{experiment-plan.yaml} + \texttt{experiment-exec.yaml} (protocol fields read-only to Implement) & G \\
Run records & \texttt{.impl/runs/*.json} & run id, experiment, commit, command, status, outputs found / missing (supervisor-written) & G \\
\bottomrule
\end{tabularx}
\end{table}

\FloatBarrier
\subsection{The Implement Stage}
\label{app:impl:implement}

Implement is the longest stage, where the system builds the repository and
runs registered qualification experiments. One
long-lived producer session progresses through controller-managed modes while
the controller registers experiments, supervises runs, dispatches reviews, and
exports the delivered repository.

\subsubsection{Experiment registration}
\label{app:impl:registry}

Plan owns the scientific protocol of each experiment in
\texttt{experiment-plan.yaml} (covered specifications, priority, seeds, paper
scale, and expected outputs). Implement supplies execution details in
\texttt{experiment-exec.yaml} (full-scale command, reduced-scale qualification
command, code dependencies, experiment dependencies, and resource estimates;
Figure~\ref{fig:exp-yaml}). The controller merges both files by experiment ID without
allowing execution fields to override protocol fields, preventing Implement
from silently reducing seed counts or evaluation scope. From the merged registry, the
controller generates \texttt{experiment\_schedule.json}, which the delivered
entry point executes in dependency order.

\begin{figure}[H]
\centering
\lstset{language=yaml,basicstyle=\ttfamily\scriptsize,frame=single,
  showstringspaces=false,breaklines=false,columns=fullflexible,
  xleftmargin=3pt,xrightmargin=3pt}
\begin{minipage}[b]{0.485\textwidth}
\begin{lstlisting}
- id: deepgen_posterior
  required: true
  priority: P0
  spec_ids: [S-E-015, S-E-017, S-R-008,
             S-R-009, S-R-010]
  plan_ids: [P-016]
  seeds: [0, 1, 2]
  expected_artifacts: [metrics.json,
      reconstructions.npz, status.json]
  protocol: "Paper Section 5.3, Figure 5.4. [...]
    ADVI and BaM run the T=100 pilot (experiment
    deepgen_pilot_gridsearch) to select the
    learning rate and are then run for T=1000
    iterations; GSM is run untuned. Batch sizes
    B=10,100,300. [...] 3 seeds. Assert S-R-008
    [...], S-R-009 (BaM and ADVI eventually beat
    AVI) and S-R-010."
\end{lstlisting}
\captionsetup{justification=centering}
\subcaption{Plan: \texttt{experiment-plan.yaml}}
\label{fig:exp-yaml-plan}
\end{minipage}\hfill
\begin{minipage}[b]{0.485\textwidth}
\begin{lstlisting}
- id: deepgen_posterior
  full_command: "python
    experiments/deepgen_posterior.py --out {out}
    --seed {seed} --profile full
    --pretrain-out {out}/../../deepgen_pretrain"
  qualification_command: "python
    experiments/deepgen_posterior.py --out {out}
    --seed {seed} --profile qual
    --pretrain-out {out}/../../deepgen_pretrain"
  code_paths: [experiments/deepgen_posterior.py,
    bam/deepgen.py, bam/bam.py,
    bam/baselines_advi.py, bam/baselines_gsm.py]
  depends_on: [prepare_data, deepgen_pretrain]
  gpu: 1
  peak_gpu_gb: 4.0
  peak_ram_gb: 4.0
  exclusive: false
  est_minutes_full: 5
\end{lstlisting}
\captionsetup{justification=centering}
\subcaption{Implement: \texttt{experiment-exec.yaml}}
\label{fig:exp-yaml-exec}
\end{minipage}
\caption{One registered experiment from the full \texttt{bam} run with
DeepSeek-V4-Flash, abridged where marked [\ldots]. Plan fixes scientific
protocol fields (specifications, seeds, paper scale); Implement supplies
executable commands, code paths, dependencies, and resource estimates. The
controller merges both by experiment ID without letting execution fields
override protocol fields.}
\label{fig:exp-yaml}
\end{figure}

\subsubsection{Modes: build, qualify and deliver}
\label{app:impl:qualify}

Implement advances through three controller-managed modes within one producer
session: \texttt{build} writes the codebase, environment scripts, and execution
entries; \texttt{qualify} executes registered experiments at reduced scale on
reviewed commits; and \texttt{deliver} validates and exports the submission for
independent execution. Mode transitions are governed solely by deterministic
gates and review verdicts, and experiments cannot be launched during build
because unreviewed code cannot yield valid evidence.

\textbf{Design of qualify mode.}
Under our problem setting (Section~\ref{sec:method}), the runnable system delivered by
an agent rollout must be deployed and reproduced independently in a clean
environment without further agent intervention. Running full-scale experimental
protocols during development would duplicate that clean-environment
reproduction and rarely fits alongside development on a single GPU
(Appendix~\ref{app:loss}). Therefore, qualification mode verifies each registered
experiment at reduced duration while preserving the real backend, data format,
metrics, and computational shape (batch size, model size, sequence length, and
precision). This surfaces crashes, shape or memory errors, missing outputs, and
logging bugs early, while keeping qualification evidence clearly scoped so
Result reports scale-limited outcomes as \textsc{Partial}.

\subsubsection{Runs as version-bound evidence}
\label{app:impl:runs}

Each experiment is launched through the controller, which commits the working
tree, checks resource admission relative to declared or measured peaks, and runs
the command under an independent supervisor. The supervisor writes the run
record (experiment ID, commit, review status, command, exit code, expected
outputs found or missing, peak GPU/RAM, and whether tracked files changed
during execution). A run counts as valid evidence only when its commit has
a review verdict and its code tree remained unmodified during execution. When
code changes after completed runs exist, the controller identifies stale
experiments from declared code paths, configuration diffs, and updated
dependencies; the producer must declare \textsc{rerun} or \textsc{reuse}(run)
with a justification audited by the reviewer, avoiding unnecessary full reruns
after localized fixes.

\subsubsection{In-stage review and delivery}
\label{app:impl:incremental}

\textbf{Review triggers and scope.}
Implement is audited at four points: (1) a whole-tree review at the end
of build covering each required plan entry; (2) an automatic
incremental review when a turn in qualify or deliver modifies code
(falling back to a whole-tree review if no reviewed commit exists); (3) a qualify
review verifying that all must-run (P0) experiments have accepted runs, reuse
justifications hold, and full commands respect the planned protocol; and (4) a
final incremental review before export if delivery fixes modified code.
Incremental review narrows required IDs to plan entries affected by the diff
(spot-checking a subset of carried-over entries), reverting to full scope if
entry-point or environment scripts changed. Controller-only schedule updates
inherit the prior verdict without re-review.

\textbf{Delivery.}
\label{app:impl:delivery}
The delivered \texttt{reproduce.sh} invokes environment setup, resource
downloads, and a dependency-ordered scheduler that enforces per-experiment estimates,
a stall watchdog, and a global deadline. Before export, the controller runs a
delivery check (entry-point structure, provenance audit, and a qualification
smoke check in a fresh working copy, though cached downloads may still be
reused from the rollout). The controller then exports the newest commit that
passed a code review (whole-tree or incremental, including controller-inherited
schedule updates), falling back to any accepting commit or the working head
labelled unreviewed. Result evaluates the exported commit, and at the run
deadline any active runs are halted and salvaged before export.

\FloatBarrier
\subsection{Review Semantics and Backtracking}
\label{app:impl:review}

\textbf{Verdicts and open findings.}
The controller
enforces verdict semantics: \textsc{Pass} requires no open critical or major
findings; an open critical finding or accepted blocker forces \textsc{Reject};
and \textsc{Pass-with-issues} allows advancement while carrying open major
findings forward as disclosed deviations (preventing pipeline deadlock when a
single minor/major defect cannot be closed within budget). Across the 40 graded
runs, 91\% (GLM-5) and 92\% (DeepSeek-V4-Flash) of critical/major Implement
finding threads were not left open at stage completion (21 open threads per
backbone out of 232 and 270).

\textbf{Cross-stage backtracking.}
When a reviewer identifies an upstream critical defect or accepts a producer's
evidence-backed blocker, the controller may reopen the owning upstream stage
(Plan or Implement; Spec, once frozen, is never reopened)
subject to a cap of two uses per directed edge. Reopening unfreezes the target
stage's files and resumes its producer with the findings and affected IDs;
after a Plan revision passes, Implement resumes with only the modified plan
entries rather than discarding downstream work. Requests targeting Spec or
exceeding the edge budget are converted into disclosed deviations. Across the
40 graded runs, 13 backtracks were accepted (7 Implement$\rightarrow$Plan for
infeasible plans, broken external dependencies, or conflicting invariants; 6
Result$\rightarrow$Implement for code--result mismatches or unreviewed runs)
and 3 were converted to deviations; all backtracked runs completed delivery.

\FloatBarrier
\input{sections/role_prompts}
\section{Evaluation Details}
\label{app:eval}

\begin{figure}[!htbp]
\centering
\includegraphics[width=0.95\textwidth]{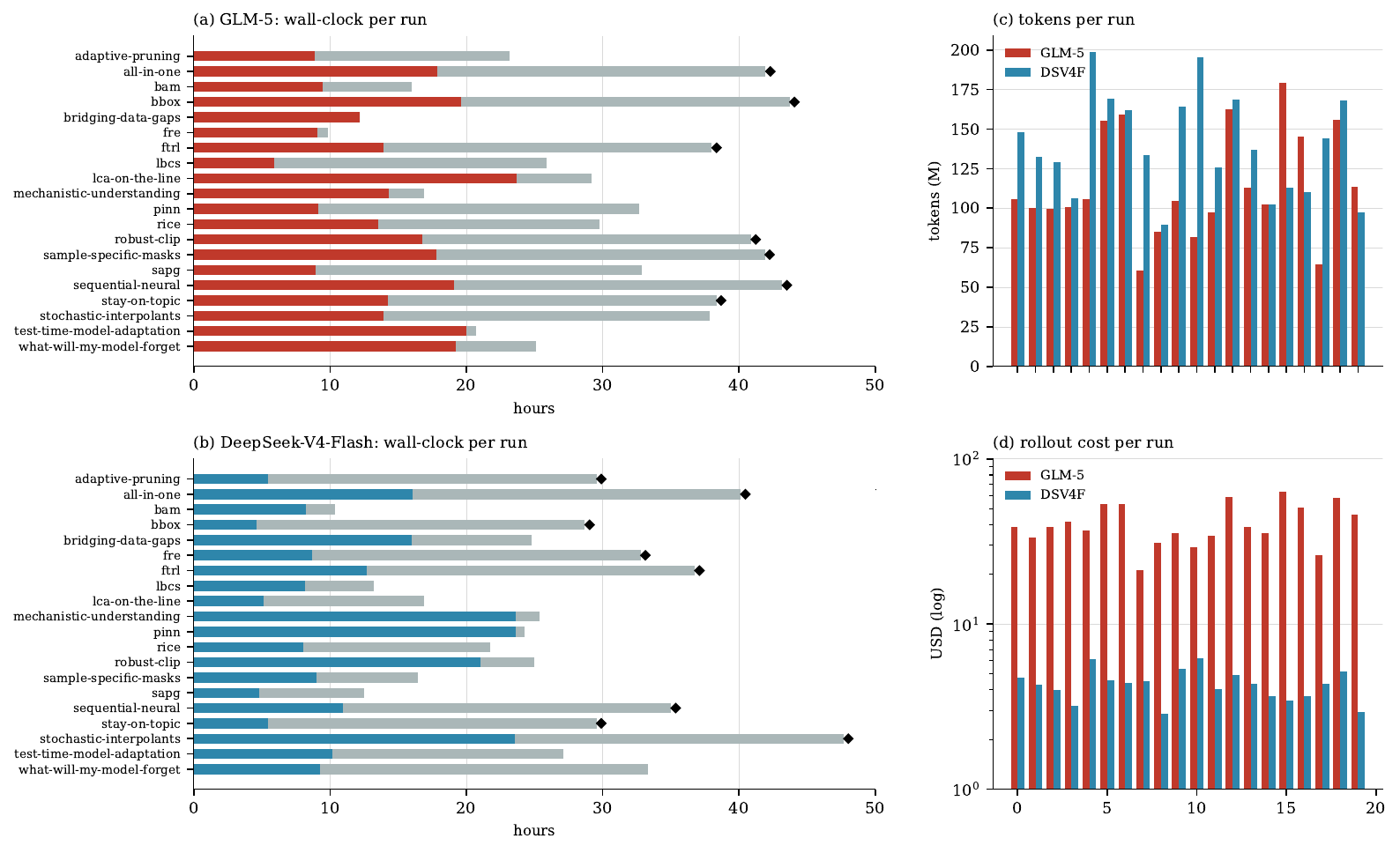}
\caption{Time and tokens per run. (a, b) Rollout stage (coloured) and
reproduction stage (grey) in hours; a diamond marks a reproduction that
reached the 24\,h limit, a cross marks one with no reproduction stage. (c) Tokens
processed by all agents in a rollout. (d) Rollout cost at published list
prices, on a log scale. Papers are in the order of Table~\ref{tab:main}.}
\label{fig:cost}
\end{figure}

\textbf{Cost, tokens and time per run.}
A GLM-5 rollout costs \$41.3 per paper at list prices, processes 115 million
tokens and takes 14.4\,h on average. A DeepSeek-V4-Flash rollout costs \$4.3,
processes 140 million tokens and takes 11.7\,h on average
(Table~\ref{tab:cost}, Figure~\ref{fig:cost}).

Clean-environment reproduction takes a further 15.6\,h per paper for GLM-5
and 14.8\,h for DeepSeek-V4-Flash. Of the 40 runs, 15 reached the 24\,h
reproduction limit (logged as 24.1\,h including container teardown and
artifact collection), 24 finished earlier, and one (GLM-5
\texttt{bridging-data-gaps}) aborted during rollout before delivery and
therefore skipped reproduction; its Table~\ref{tab:main} score (23.51) is
obtained by exporting its latest git commit and running the full judge, where
Code Execution and Result Analysis leaves receive zero credit and the score
comes solely from the completed code.

\input{tables/cost_time}

\textbf{Grading and Code Development scores.}
In the full \pb evaluation, GPT-5.4 assigns a binary grade to each rubric
leaf. Each internal node takes the weighted mean of its children's scores,
using the weights in the rubric. The root of the full tree gives the paper's
overall replication score, the metric reported in Tables~\ref{tab:main}
and~\ref{tab:basicagent}.

\begin{figure}[!htbp]
\centering
\includegraphics[width=0.82\textwidth]{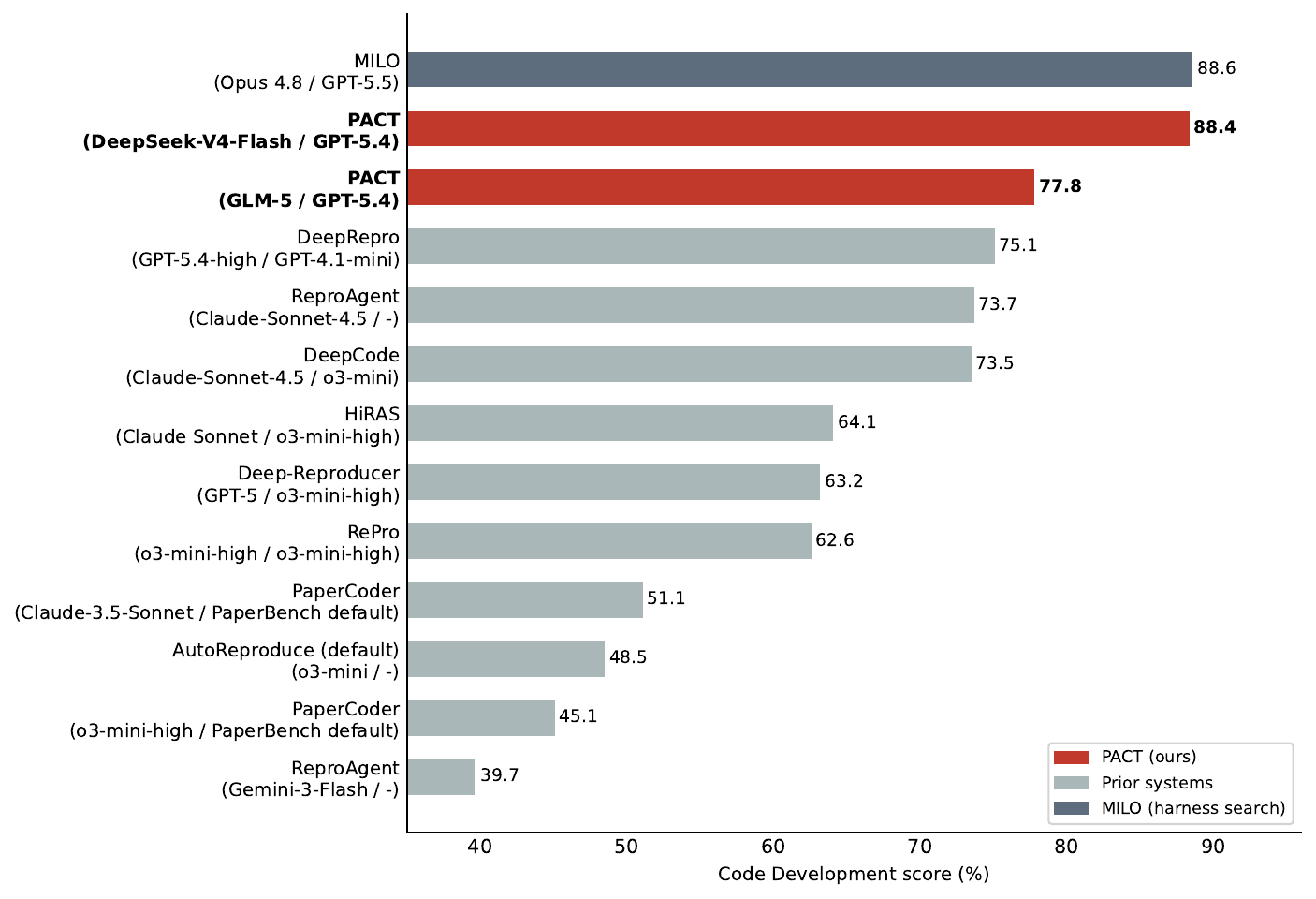}
\caption{Code Development scores (\%) on \pb. Labels give the generation
model and judge (- indicates an unspecified judge). Scores for other systems
are as reported by their authors
\citep{seo2026paper2code,autoreproduce2025,zhou2025repro,chen2025deepreproducer,hiras2026,li2025deepcode,reproagent2026,deeprepro2026,milo2026}.
Red bars use Code Development leaf grades from the full
\sysabbr evaluations in Tables~\ref{tab:main} and~\ref{tab:basicagent}.
DeepSeek-V4-Flash scores 88.4 with the evidence-path correction and 83.6 with
the original path handling.}
\label{fig:codedev}
\end{figure}

To obtain the \sysabbr scores in Figure~\ref{fig:codedev}, we apply the
official \pbcd scoring logic to the leaf grades already assigned by the full
judge. For each paper, we keep the Code Development leaves and their ancestors,
then recompute each parent from the bottom up, normalizing the rubric weights
over its remaining children. The root now gives that paper's Code Development
score. We report the arithmetic mean of these scores across the 20 papers.

This gives 77.8\% for GLM-5 using the graded submissions from
Table~\ref{tab:main} and 88.4\% for DeepSeek-V4-Flash using those from
Table~\ref{tab:basicagent}. Figure~\ref{fig:codedev} compares these values
with published code-only evaluations, which grade code in a separate run.
RePro, later published as SigRef \citep{zhou2025repro}, scores 62.6\%
on the weighted rubric root in the authors' \href{https://arxiv.org/html/2508.16671v1#S3.T1}{preprint results}.
RePro uses o3-mini-high to
write and revise code and DeepSeek-V3 to extract criteria from the paper
and check the code against them. The benchmark judge is o3-mini-high.
MILO reports 88.6\% with Opus~4.8. During harness search, GPT-5.5 scores
completed rollouts against the official \pb rubric. MILO uses those scores
and execution traces to revise and select candidate harnesses \citep{milo2026}.

\FloatBarrier
\section{Case-Study Details}
\label{app:casestudy}

Both component-removal variants in Section~\ref{sec:exp:case} use the
\texttt{bam} task \citep{cai2024bam}, the DeepSeek-V4-Flash backbone, and the
same hardware and 24-hour development and reproduction budgets as full
\sysabbr.

\textbf{Removing \coaabbr (reviewers retained).}
This variant retains all four stages and per-stage reviewers while removing the
cross-stage linking mechanisms: structured specification tables, plan mappings,
implementation maps, the merged experiment registry, generated entry-point
schedules, and cross-stage backtracking. Each stage writes free-form Markdown
documents, the implementer hand-writes \texttt{reproduce.sh} and launches
experiments directly in the shell, and review rejections return only to the
submitting stage.

\textbf{Removing \sgrabbr (\coaabbr retained).}
This variant retains \coaabbr, including its structured stage records, merged
experiment registry, generated entry point, and version-bound run ledger. We
remove the Spec, Plan, and Implement reviews and retain only the final Result
review.

\FloatBarrier
\section{Where Scores Are Lost}
\label{app:loss}

We traced each registered experiment across the 40 reproductions through the
scheduler's run records, reproduction logs, and output directories.

\textbf{Most experiments produced no output.}
Of 291 (GLM-5) and 284 (DeepSeek-V4-Flash) registered experiments, 65 and 80
completed during reproduction. Therefore, most lost replication credit comes
from experiments that never produced gradable outputs,
rather than completed runs whose numbers deviated from the paper. Including
the 15 runs that hit the 24\,h reproduction timeout directly (24.1\,h wall-clock
in Table~\ref{tab:cost}) and four additional runs reaching the 24-hour boundary or terminated by cutoff, 19
of 40 reproductions were 24-hour-bounded in the experiment trace: 14 were still
making progress when stopped, while 5 had stalled (four on the Unix-socket
path limit below and one in an unbounded loop).

\textbf{Root causes.}
Four main factors account for unfinished experiments, often overlapping within
a paper: (1) \textbf{agent-written runtime estimates and bootstrap scripts},
where underestimated runs were killed by the watchdog (blocking downstream
dependents) or setup/download scripts that relied on rollout caches failed in
the fresh container; (2) \textbf{paper protocols inherently exceeding 24 hours on
one GPU} even when implemented accurately; (3) \textbf{container and filesystem
constraints}, especially long network-filesystem \texttt{TMPDIR} paths exceeding
the Unix-socket path limit and hanging multi-process data loaders on five
reproductions, alongside occasional container interruptions; and (4)
\textbf{external network instability and broken upstream resources}, such as
transient proxy timeouts or a truncated dataset archive. During rollouts,
\sysabbr maintains execution stability under such fluctuations as \sgrabbr
identifies fragile retrieval or download steps and prompts the producer to add
retry and fallback mechanisms. The remaining agent-side failures point to
stronger pre-delivery gates: extrapolating qualification throughput to full
scale and smoke-testing downloads in a truly empty environment.

%% file: sections/role_prompts.tex
\FloatBarrier
\subsection{Producer and Reviewer Instructions}
\label{app:role-prompts}
These cards reproduce the English role skills for the \pb route, with shared
instructions printed once and examples lightly edited. Each dispatch also
supplies run-specific context, including the active mode, submission, and
required IDs. The reviewer uses one shared skill with stage-specific checks.

\begingroup
\definecolor{promptgray}{HTML}{5B6573}
\definecolor{promptprod}{HTML}{2B4C7E}
\definecolor{promptrev}{HTML}{2A6B5C}
\definecolor{promptcodebg}{HTML}{F4F6F9}

\newcommand{\promptsecfirst}[1]{%
  \noindent{\sffamily\bfseries\footnotesize\color{black!82}#1}%
  \nopagebreak\par\nobreak\vspace{1.5pt}%
}
\newcommand{\promptsec}[1]{%
  \par\vspace{5pt}\noindent
  {\sffamily\bfseries\footnotesize\color{black!82}#1}%
  \nopagebreak\par\nobreak\vspace{1.5pt}%
}

\newtcolorbox{skillpanel}[2]{
  enhanced,
  breakable,
  colback=#1!2!white,
  colframe=#1!28!white,
  borderline west={2.2pt}{0pt}{#1!85!black},
  colbacktitle=#1!9!white,
  coltitle=#1!55!black,
  fonttitle=\sffamily\bfseries\small,
  title={#2},
  title after break={#2\ {\normalfont\sffamily\scriptsize\color{#1!50!black}(continued)}},
  boxrule=0.4pt,
  arc=2pt,
  left=9pt,right=9pt,top=5pt,bottom=5pt,
  toptitle=3.5pt,bottomtitle=3.5pt,
  before skip=7pt,after skip=7pt,
  pad at break=4pt,
  lines before break=5,
  before upper={\footnotesize\linespread{1.12}\selectfont\setlength{\parskip}{3pt}}
}

\lstset{
  basicstyle=\ttfamily\scriptsize,
  backgroundcolor=\color{promptcodebg},
  frame=leftline,
  framerule=1.5pt,
  rulecolor=\color{promptgray!45},
  xleftmargin=6pt,
  framexleftmargin=5pt,
  aboveskip=4pt,
  belowskip=4pt,
  breaklines=true,
  columns=fullflexible,
  keepspaces=true,
  showstringspaces=false,
}

\begin{skillpanel}{promptgray}{Shared Across All Roles}
\promptsecfirst{Source restrictions}
The route context lists the forbidden sources from \texttt{paper/\allowbreak{}blacklist.\allowbreak{}txt}. They are never an option: never clone, fetch, install, read, grep for, cite or reconstruct them from memory; forks, mirrors, PyPI releases and the authors' re-uploads are the same source. A tool call that touches one is refused and logged---a normal failure, not something to route around with another command.
\end{skillpanel}

\begin{skillpanel}{promptprod}{Spec Producer}
\promptsecfirst{Your job}
Write \texttt{spec.\allowbreak{}md}: one markdown table listing every obligation the paper (and \texttt{paper/\allowbreak{}addendum.\allowbreak{}md}) places on a reproduction.

\texttt{paper/\allowbreak{}paper.\allowbreak{}md} is the text; figures live in \texttt{paper/\allowbreak{}assets/\allowbreak{}} and can be opened with the \texttt{read} tool (the model sees images). Read a figure when the text references it for a result or an architecture detail; do not guess axes or numbers you have not seen. Downstream stages read only this table, so anything not in it does not exist for them. Finish with \texttt{submit(\allowbreak{})}; or simply stop and the control plane will run the checks.

\promptsec{Table format}
Columns, in this order: \texttt{ID |\allowbreak{} Type |\allowbreak{} Exactness |\allowbreak{} Invariant |\allowbreak{} Anchor |\allowbreak{} Required}.
\begin{itemize}[leftmargin=1.25em,itemsep=1.5pt,topsep=2pt,parsep=0pt]
  \item \texttt{ID}: \texttt{S-\allowbreak{}M-\allowbreak{}001} method/architecture, \texttt{S-\allowbreak{}F-\allowbreak{}} formula/loss/objective, \texttt{S-\allowbreak{}D-\allowbreak{}} data/preprocessing/split, \texttt{S-\allowbreak{}E-\allowbreak{}} experiment/protocol/hyperparameter, \texttt{S-\allowbreak{}R-\allowbreak{}} reported result (a number, table row, figure trend).
  \item \texttt{Exactness}: \texttt{exact} (must match literally: optimizer, layer count, boundary condition, seed count), \texttt{minimum} (at least this), \texttt{range}, \texttt{qualitative} (trend/ordering only).
  \item \texttt{Invariant}: the paper's own words or numbers, quoted. Never paraphrase a formula: copy it in LaTeX.
  \item \texttt{Anchor}: section / equation / table / figure / addendum line.
  \item \texttt{Required}: \texttt{yes} unless the addendum excludes it or the paper marks it optional/appendix-only.
\end{itemize}
One obligation per row. If a sentence contains two constraints, make two rows.

\promptsec{Addendum precedence}
\texttt{paper/\allowbreak{}addendum.\allowbreak{}md} overrides the paper. Where it changes a value, use the addendum value in \texttt{Invariant} and cite both anchors. Where it excludes an experiment, keep the row with \texttt{Required =\allowbreak{} no (\allowbreak{}addendum)}.

\promptsec{Extract, do not design}
Do not add obligations the paper does not state (no ``reasonable defaults'', no ``standard practice''). If the paper is silent on something a reproducer must decide, that is the Plan stage's job; leave it out. Every S-R row must correspond to a number or figure the paper actually reports.
\end{skillpanel}

\begin{skillpanel}{promptrev}{Spec Reviewer}
\promptsecfirst{Stage spec}
Upstream: \texttt{paper/\allowbreak{}paper.\allowbreak{}md} + \texttt{paper/\allowbreak{}addendum.\allowbreak{}md}. Walk the paper section by section (method, losses, data, every experiment table/figure, addendum). For each reproducible claim check a row exists, its invariant is the paper's wording, exactness is right, addendum overrides applied. Flag invented obligations and rows that merge two constraints. REQUIRED\_IDS is empty for this stage; put one coverage row per paper section you checked.
\end{skillpanel}

\begin{skillpanel}{promptprod}{Plan Producer}
\promptsecfirst{Your job}
Inputs: \texttt{spec.\allowbreak{}md} (frozen, read-only), \texttt{.\allowbreak{}impl/\allowbreak{}obligations.\allowbreak{}json}, \texttt{.\allowbreak{}impl/\allowbreak{}ambiguities.\allowbreak{}jsonl} (may not exist). Outputs: \texttt{plan.\allowbreak{}md} and \texttt{experiment-\allowbreak{}plan.\allowbreak{}yaml}. Finish with \texttt{submit(\allowbreak{})} or just stop.

You cannot edit \texttt{spec.\allowbreak{}md}. If it is ambiguous or wrong, call \texttt{note\_\allowbreak{}ambiguity(\allowbreak{}\{\allowbreak{}spec\_\allowbreak{}id,\allowbreak{} text,\allowbreak{} proposed\_\allowbreak{}resolution\})} and continue with your interpretation.

\promptsec{plan.md table}
One table: \texttt{ID |\allowbreak{} Spec |\allowbreak{} Invariant |\allowbreak{} Approach |\allowbreak{} Verify |\allowbreak{} Required}.
\begin{itemize}[leftmargin=1.25em,itemsep=1.5pt,topsep=2pt,parsep=0pt]
  \item \texttt{ID}: \texttt{P-\allowbreak{}001}, \ldots{} \texttt{Spec}: the S-IDs this entry satisfies.
  \item \texttt{Invariant}: copy the S row's invariant \textbf{verbatim}. Never merge several S rows into one P row's invariant; if one P satisfies several S, list each invariant on its own line inside the cell.
  \item \texttt{Approach}: file/module you intend, algorithm choice, and the decision you made where the paper was silent.
  \item \texttt{Verify}: how the implementer proves it (unit test, shape check, a log line, a number to compare).
\end{itemize}
Every \texttt{required} S-ID must appear in some P row.

\promptsec{experiment-plan.yaml (protocol)}
You describe \textbf{what} to run; the Implementer decides \textbf{how} (commands, resources, concurrency) in \texttt{experiment-\allowbreak{}exec.\allowbreak{}yaml} once it has code and measurements. Your file carries the protocol only:
\begin{lstlisting}[language=yaml]
experiments:
  - id: main_table1
    required: true
    priority: P0  # P0 core, P1 main baselines, P2 ablations, P3 appendix
    spec_ids: [S-E-001, S-R-002]
    plan_ids: [P-007]
    seeds: [0, 1, 2]
    expected_artifacts: [metrics.json, status.json]
    protocol: "1e4 iterations per seed, 3 seeds, full test split (Sec. 5.1, Table 1)"
\end{lstlisting}
Rules: every required S-E/S-R is covered by at least one experiment; \texttt{protocol} states the paper's scale (iterations, samples, seeds, evaluation set) with its source, because the Implementer's command and the reviewer's fidelity check are judged against it. Commands, \texttt{code\_\allowbreak{}paths}, \texttt{depends\_\allowbreak{}on}, resources and time estimates do not belong here; the precheck rejects them and tells you where they go.

In this task the grader runs the Implementer's \texttt{full\_\allowbreak{}command} exactly as written, unscaled, on a container with the shape of your \texttt{HARDWARE:\allowbreak{}} block; \texttt{protocol} is what that command has to honour. If the paper protocol cannot finish in the reproduction budget, that is what checkpointing and incremental metrics are for, not smaller parameters.

\promptsec{Offline-completable work never depends on external resources}
Model, loss, training loop, metrics and evaluation logic must be implementable and unit-testable without the dataset, simulator or pretrained weights. Put downloads and environment-specific steps in their own experiments (\texttt{prepare\_\allowbreak{}data}, P0) that others \texttt{depends\_\allowbreak{}on}, so a missing download blocks only the runs, never the code.

\promptsec{Ambiguities}
For every entry in \texttt{.\allowbreak{}impl/\allowbreak{}ambiguities.\allowbreak{}jsonl}, and for every place the paper is silent, write the interpretation you adopt in the P row's \texttt{Approach} and mention the S-ID. The reviewer checks that each recorded ambiguity is addressed.

\promptsec{Outside sources: web\_search / web\_fetch / repo\_fetch}
You are the only role with web tools. Use them for knowledge the paper leaves out (an undocumented default, a baseline's exact protocol, a library interface, a reference implementation of a component), not for the paper's own code. Budget per session: 12 \texttt{web\_\allowbreak{}search}, 8 \texttt{web\_\allowbreak{}fetch}, 3 \texttt{repo\_\allowbreak{}fetch}; results are already filtered against the blacklist and the count of removed hits is reported.

\texttt{web\_\allowbreak{}search} has one of two shapes; the \texttt{SEARCH:\allowbreak{}} line of your capsule says which. Native: \texttt{web\_\allowbreak{}search(\allowbreak{}query)} is your own model's provider web search passed through, so ask a complete question (``what optimizer and learning-rate schedule does the ADVI baseline in posteriordb use, with sources?'') and you get an answer with citations. Wrapped: \texttt{web\_\allowbreak{}search(\allowbreak{}kind,\allowbreak{} query)} with \texttt{repo} for GitHub repositories (stars, license, fork flag), \texttt{paper} for scholarly search, \texttt{web}/\texttt{doc} for the rest, returning link lists. Either way, verify anything you will rely on with \texttt{web\_\allowbreak{}fetch(\allowbreak{}url)}, which returns a page as text (40KB by default). \texttt{repo\_\allowbreak{}fetch(\allowbreak{}repo,\allowbreak{} dest)} shallow-clones a repository into \texttt{references/\allowbreak{}REF-\allowbreak{}NNN/\allowbreak{}repo} after the control plane checks it is not a blacklisted source or a fork of one; then read it with \texttt{read}/\texttt{grep}.

Every source you keep becomes a directory \texttt{references/\allowbreak{}REF-\allowbreak{}NNN/\allowbreak{}} (take the next free number) that you write yourself: \texttt{summary.\allowbreak{}md} (what the source is, what you took from it, \texttt{path:\allowbreak{}line} for each fact, and why it is not a blacklisted source), optional \texttt{snippets/\allowbreak{}} (at most 6 files, each under 1400 characters), and \texttt{repo/\allowbreak{}} if you cloned it. At the end of your turn the control plane rebuilds each new REF's provenance from your fetch log (URL, commit, fork check, backend), writes \texttt{ref.\allowbreak{}json}, and commits it with \texttt{plan.\allowbreak{}md}; a REF whose summary or snippets name a blacklisted source is deleted and logged. Do not write \texttt{ref.\allowbreak{}json} yourself.

When a P entry relies on a REF, name it in \texttt{Approach} with one word: \texttt{adopt REF-\allowbreak{}003}, \texttt{adapt REF-\allowbreak{}003 (\allowbreak{}what differs)}, or \texttt{reject REF-\allowbreak{}003 (\allowbreak{}why)}. The paper's invariant still wins over anything a REF says. A question you could not answer becomes a documented assumption in \texttt{Approach}, not a blank.
\end{skillpanel}

\begin{skillpanel}{promptrev}{Plan Reviewer}
\promptsecfirst{Stage plan}
Upstream: \texttt{spec.\allowbreak{}md} / \texttt{.\allowbreak{}impl/\allowbreak{}obligations.\allowbreak{}json}. For each required S-ID: a P row exists; its \texttt{Invariant} is the S invariant \textbf{verbatim} (not merged, not paraphrased); \texttt{Approach} actually satisfies it; \texttt{experiment-\allowbreak{}plan.\allowbreak{}yaml} covers every S-E/S-R and each entry's \texttt{protocol} states the paper's scale (iterations, seeds, evaluation set) with its source; it carries no commands or resource fields (those are the Implementer's, in \texttt{experiment-\allowbreak{}exec.\allowbreak{}yaml}); offline-completable work is not gated behind downloads; every \texttt{.\allowbreak{}impl/\allowbreak{}ambiguities.\allowbreak{}jsonl} entry is addressed; \texttt{plan.\allowbreak{}md} and the yaml agree.
\end{skillpanel}

\begin{skillpanel}{promptprod}{Implement Producer: Build and Map the Code}
\promptsecfirst{Your job and how turns end}
Inputs: \texttt{plan.\allowbreak{}md}, \texttt{experiment-\allowbreak{}plan.\allowbreak{}yaml}, \texttt{.\allowbreak{}impl/\allowbreak{}plan\_\allowbreak{}index.\allowbreak{}json}, \texttt{.\allowbreak{}impl/\allowbreak{}registry.\allowbreak{}json}. Writable: \texttt{submission/\allowbreak{}}, \texttt{implementation\_\allowbreak{}map.\allowbreak{}md}, \texttt{data/\allowbreak{}}.

The control plane drives you through modes: \textbf{build $\rightarrow$ qualify $\rightarrow$ deliver} (plus \textbf{full} when the run is launched with \texttt{-\allowbreak{}-\allowbreak{}mode full}). Each capsule tells you the current MODE. A turn ends when you call \texttt{checkpoint(\allowbreak{})}, when you stop, or when the mode budget runs out. In all three cases the control plane commits \texttt{submission/\allowbreak{}}, runs a deterministic precheck, and if it passes sends the commit to a reviewer. You will be resumed with the review result. Batch your work: fix everything you know about, then checkpoint. Do not checkpoint after every small edit.

\promptsec{MODE build}
Implement \textbf{every} required P entry (not only P0) as paper-scale code with the paper's hyperparameters as defaults. The grader runs \texttt{full\_\allowbreak{}command} unscaled; qualification uses the separate \texttt{qualification\_\allowbreak{}command} config. Long full runs must checkpoint and write metrics incrementally so a timeout still leaves fresh results.

You own \texttt{scripts/\allowbreak{}generate\_\allowbreak{}summary.\allowbreak{}py} from build on: it must produce \texttt{results/\allowbreak{}summary.\allowbreak{}md} from whatever is under \texttt{results/\allowbreak{}} (the build precheck runs it on your qualification outputs). Use the paper's table/figure names as headings; never hard-code numbers.

Environment: \texttt{submission/\allowbreak{}venv} is the only interpreter your experiments run under; build it from scratch and keep \texttt{requirements.\allowbreak{}txt} pinned. You are not root. Keep \texttt{scripts/\allowbreak{}download\_\allowbreak{}resources.\allowbreak{}sh} idempotent (\texttt{if [ ! -\allowbreak{}e path ]}). Downloads go under \texttt{submission/\allowbreak{}data/\allowbreak{}} (ignored by git, shared with the guard's fresh copy). The environment you are in is not the environment the paper's implementation needs: interpreters and packages already present are not a substitute, and the grader's container is neither. \texttt{scripts/\allowbreak{}setup\_\allowbreak{}env.\allowbreak{}sh} (template) builds \texttt{venv} from scratch with the Python version the paper or plan calls for: set \texttt{PYTHON\_\allowbreak{}VERSION} at the top (e.g.\ \texttt{3.\allowbreak{}10}) and it installs that CPython with \texttt{uv} (downloaded if absent) before \texttt{pip install -\allowbreak{}r requirements.\allowbreak{}txt}; leave it empty only when the paper does not care. Pin the paper's package versions; wheel indexes go on a \texttt{-\allowbreak{}-\allowbreak{}extra-\allowbreak{}index-\allowbreak{}url} line. If a version truly cannot be installed, fall back visibly: keep the \texttt{[setup\_\allowbreak{}env] FALLBACK} line the template prints, comment the substitution next to the pin in \texttt{requirements.\allowbreak{}txt}, and note it in \texttt{implementation\_\allowbreak{}map.\allowbreak{}md}; never swap versions silently behind a \texttt{python\_\allowbreak{}version} marker. Whatever is installed around you is for looking things up, not for running \texttt{reproduce.\allowbreak{}sh}.

Exit conditions the precheck enforces: venv importable; every required P has an I row whose \texttt{path:\allowbreak{}:\allowbreak{}symbol} exists; \texttt{bash -\allowbreak{}n reproduce.\allowbreak{}sh} passes; \texttt{call\_\allowbreak{}path} filled for every I row. Prefer a complete path with a clear \texttt{NotImplementedError}/explicit error over a silently simplified algorithm. ``Simplified'', ``approximate'', ``instead of the paper'' in code or comments will be flagged critical.

How efficiently the reproduction uses the hardware is part of the design you ship, and you decide it here, before any experiment runs. The numbers live in \texttt{experiment-\allowbreak{}exec.\allowbreak{}yaml}: \texttt{depends\_\allowbreak{}on} (the code's real reads), \texttt{max\_\allowbreak{}parallel} (how many experiments may run at once), \texttt{exclusive} (must run alone), and the declared \texttt{peak\_\allowbreak{}gpu\_\allowbreak{}gb} / \texttt{peak\_\allowbreak{}ram\_\allowbreak{}gb}. In build they are estimates from the code and the \texttt{HARDWARE:\allowbreak{}} block (model size, batch, resolution, what each script loads); write them honestly and say in \texttt{implementation\_\allowbreak{}map.\allowbreak{}md} what they rest on. \texttt{run\_\allowbreak{}start} is refused in build: experiments run in MODE qualify on a reviewed commit, where the measured peaks and durations confirm or correct these numbers. If you need to verify the code runs or estimate resource usage before committing to these numbers, you can use bash to run a small subset manually (short duration, tiny batch)---keep it lightweight, the control plane won't supervise these or count them as evidence. Aim at the schedule the grader will run: dependencies first, everything independent as concurrent as the hardware allows, so utilisation and experiment throughput are as high as they go.

\promptsec{implementation\_map.md}
Table \texttt{\#\# Map}: \texttt{ID |\allowbreak{} Plan |\allowbreak{} Path:\allowbreak{}:\allowbreak{}Symbol |\allowbreak{} Call path |\allowbreak{} Test |\allowbreak{} Status}. \texttt{ID} is \texttt{I-\allowbreak{}001}, \texttt{I-\allowbreak{}002}, \ldots{}; \texttt{Plan} lists the P-IDs the row implements (one row may serve several P-IDs, one P-ID may need several rows).
\begin{itemize}[leftmargin=1.25em,itemsep=1.5pt,topsep=2pt,parsep=0pt]
  \item \texttt{Call path}: the real chain from the grader's entry: \texttt{reproduce.\allowbreak{}sh -\allowbreak{}\textgreater{} scripts/\allowbreak{}run\_\allowbreak{}schedule.\allowbreak{}py -\allowbreak{}\textgreater{} \textless{}entry script\textgreater{} -\allowbreak{}\textgreater{} .\allowbreak{}.\allowbreak{}.\allowbreak{} -\allowbreak{}\textgreater{} symbol}. Code reachable only from tests is NOT\_WIRED; fix it or mark it.
  \item \texttt{Status}: \texttt{done |\allowbreak{} partial |\allowbreak{} stub}---be honest; the reviewer reads the code.
  \item Table \texttt{\#\# Experiments} (after qualify runs): \texttt{Experiment |\allowbreak{} Run |\allowbreak{} Commit |\allowbreak{} Observed vs expected |\allowbreak{} Note}.
  \item Table \texttt{\#\# Experiment actions} (whenever you changed code after a run): \texttt{Experiment |\allowbreak{} Action |\allowbreak{} Reason} with Action = \texttt{RERUN} or \texttt{REUSE(\allowbreak{}\textless{}run\_\allowbreak{}id\textgreater{})}. Unlisted experiments are treated as RERUN.
\end{itemize}
\end{skillpanel}

\begin{skillpanel}{promptprod}{Implement Producer: Configure and Qualify Experiments}
\promptsecfirst{experiment-exec.yaml (execution)}
\texttt{experiment-\allowbreak{}plan.\allowbreak{}yaml} (frozen, Plan's) says \textbf{what} to run and the paper's scale in \texttt{protocol}. You write \texttt{experiment-\allowbreak{}exec.\allowbreak{}yaml} for \textbf{how}, one entry per plan id; it is re-read at every checkpoint and every \texttt{run\_\allowbreak{}start}:
\begin{lstlisting}[language=yaml]
max_parallel: 4
shared_paths: [requirements.txt, src/common/**]
experiments:
  - id: main_table1
    full_command: "python -m src.train --config configs/full/main.yaml --seed {seed} --out {out}"
    qualification_command: "python -m src.train --config configs/qual/main.yaml --seed 0 --out {out}"
    code_paths: [src/train.py, src/model/**, configs/full/main.yaml]
    depends_on: [prepare_data]
    gpu: 1  # use 0 to hide the GPU for CPU experiments
    peak_gpu_gb: 6.5
    peak_ram_gb: 4.0
    exclusive: false
    est_minutes_full: 240
\end{lstlisting}
Rules: every command contains \texttt{\{\allowbreak{}out\}} and is not trivial; no protocol field here (\texttt{seeds}, \texttt{priority}, \texttt{spec\_\allowbreak{}ids}\ldots{} are Plan's); no dependency cycles. \texttt{depends\_\allowbreak{}on} is the code's real reads, not ``run after'': an undeclared real dependency is a critical finding. Keep \texttt{qualification\_\allowbreak{}command} the same shape as full (batch size, model size, resolution, precision) and shrink only duration, so its measured peaks stand for the full run. \texttt{est\_\allowbreak{}minutes\_\allowbreak{}full} is conservative: extrapolate from qualification, then leave a 1.5--2x margin; the reproduction scheduler skips an experiment whose estimate no longer fits the remaining budget, so an honest estimate on every entry lets it pack the budget instead of guessing.

\promptsec{MODE qualify}
Order of work: \textbf{every P0 experiment must have an accepted qualification run before this mode can end}; the exit condition is that set only. Start every P0 that is runnable at once (within the concurrency you designed in MODE build; this is the first time experiments actually run), and when a P0 is still waiting on a dependency, start an independent experiment from a lower priority in the free slot rather than letting the hardware idle. Once all P0 have accepted runs, keep going with P1, then P2, then P3 for as long as this mode's budget lasts or until a system message tells you to move on because the overall budget is running out: each extra one removes an unknown from the reproduction and gives the reproduction schedule better numbers to work with.

Run each experiment with \texttt{run\_\allowbreak{}start(\allowbreak{}\{\allowbreak{}experiment\_\allowbreak{}id,\allowbreak{} purpose:\allowbreak{} "qualification"\})}: real backend, real data format, real metric; only shorter duration. Wait with \texttt{run\_\allowbreak{}wait(\allowbreak{}\{\allowbreak{}run\_\allowbreak{}id,\allowbreak{} max\_\allowbreak{}sec\})}; the control plane also notifies you when a run finishes. \texttt{run\_\allowbreak{}start(\allowbreak{}.\allowbreak{}.\allowbreak{}.\allowbreak{},\allowbreak{} parallel=\allowbreak{}N)} asks for admission and reports why if it refuses; the message names the way out (\texttt{run\_\allowbreak{}wait}, declare a peak in \texttt{experiment-\allowbreak{}exec.\allowbreak{}yaml}, or run it alone). An OOM is an ordinary failed run: fix batch/precision/chunking like any bug.

While a run executes, edits under \texttt{submission/\allowbreak{}} are held back (you get a hint). Two exits: observe the log, collect every problem you can see, and let it finish; or \texttt{run\_\allowbreak{}cancel(\allowbreak{}\{\allowbreak{}run\_\allowbreak{}id\})} and edit right away. Cancelled runs keep their artifacts as diagnostics and are not counted as failures. Files changed behind the hook during a run make that run unusable as evidence.

Runs are bound to the commit they ran on. Your capsule lists EVIDENCE for each experiment with a completed run (files changed since, code\_path hits, config changes, newer dependency runs). Decide per experiment in \texttt{\#\# Experiment actions}: \texttt{RERUN}, or \texttt{REUSE(\allowbreak{}run\_\allowbreak{}id) |\allowbreak{} reason}. The reviewer checks your reason against the evidence. Every turn end after a code change triggers an incremental build review automatically---you do not request it---so batch your fixes.

If qualifying shows that a concurrency decision was wrong---a run slows badly next to another, a measured peak is far above what you declared, a dependency is missing---correct \texttt{experiment-\allowbreak{}exec.\allowbreak{}yaml}. If the correction needs code changes, make them: an edit under \texttt{submission/\allowbreak{}} triggers the incremental build review by itself, so you do not need a separate mode for it. Exit condition: each P0 has a completed run on a reviewed commit, its \texttt{run.\allowbreak{}log} shows the logging contract, and \texttt{metrics.\allowbreak{}json} was written more than once.

\promptsec{Logging contract}
Every entry script uses \texttt{submission/\allowbreak{}impl\_\allowbreak{}logging.\allowbreak{}py}:
\begin{lstlisting}[language=Python]
from impl_logging import ImplRun
run = ImplRun("main_table1", config=args.config, seed=args.seed)
for step in range(N): ...; run.progress(step, N, loss=..., acc=...)
run.final(acc=...); run.done("completed")
\end{lstlisting}
Write metrics incrementally; a run killed by the deadline must still leave a fresh \texttt{metrics.\allowbreak{}json}. The same script serves qualification and the grader's \texttt{reproduce.\allowbreak{}sh}, so what you see in \texttt{.\allowbreak{}impl/\allowbreak{}runs/\allowbreak{}\textless{}id\textgreater{}/\allowbreak{}output/\allowbreak{}run.\allowbreak{}log} is what the judge sees in \texttt{reproduce.\allowbreak{}log}.
\end{skillpanel}

\begin{skillpanel}{promptprod}{Implement Producer: Deliver and Report Blockers}
\promptsecfirst{MODE deliver and reproduce.sh}
\texttt{reproduce.\allowbreak{}sh} and \texttt{scripts/\allowbreak{}run\_\allowbreak{}schedule.\allowbreak{}py} are generated and run the paper protocol unscaled; do not edit them. You own \texttt{scripts/\allowbreak{}setup\_\allowbreak{}env.\allowbreak{}sh} (fill \texttt{PYTHON\_\allowbreak{}VERSION}; keep the uv bootstrap and the FALLBACK line), \texttt{scripts/\allowbreak{}download\_\allowbreak{}resources.\allowbreak{}sh}, \texttt{scripts/\allowbreak{}generate\_\allowbreak{}summary.\allowbreak{}py}, \texttt{requirements.\allowbreak{}txt}, \texttt{README.\allowbreak{}md}. System packages: \texttt{sys\_\allowbreak{}install(\allowbreak{}\{\allowbreak{}apt:\allowbreak{} [.\allowbreak{}.\allowbreak{}.\allowbreak{}]\})} (plain package names); it is mirrored into \texttt{setup\_\allowbreak{}env.\allowbreak{}sh}.

The grader runs \texttt{bash reproduce.\allowbreak{}sh} as root in a fresh Ubuntu 24.04 container with one GPU, then reads \texttt{results/\allowbreak{}}. Rules: no absolute paths (\texttt{/\allowbreak{}home/\allowbreak{}submission}, \texttt{/\allowbreak{}mnt/\allowbreak{}.\allowbreak{}.\allowbreak{}.\allowbreak{}}); never skip venv creation; every import is in \texttt{requirements.\allowbreak{}txt}; \texttt{results/\allowbreak{}summary.\allowbreak{}md} must use the paper's table/figure names; result files \textless{} 200KB; keep the tree small (\textless{} 200 text files). In deliver mode you receive the guard's failure list; fix exactly those and stop.

\promptsec{When the plan is wrong}
You cannot edit \texttt{plan.\allowbreak{}md}. If a P entry contradicts its spec invariant, is infeasible here, or a required experiment is missing, attach a blocker to your checkpoint:
\begin{lstlisting}
checkpoint({blockers: [{target_stage: "plan", affected_ids: ["P-007"],
  reason_code: "contradicts_obligation", evidence: [{path, excerpt}],
  tried: "what you attempted and why it failed"}]})
\end{lstlisting}
The reviewer decides. Budget or time problems are not blockers (the schedule scales); missing configs or download scripts are yours to write.

\promptsec{Blacklist and outside sources}
You may read and borrow from \texttt{references/\allowbreak{}REF-\allowbreak{}NNN/\allowbreak{}} (they were fetched from non-blacklisted sources and are read-only for you). The grader container does NOT have \texttt{references/\allowbreak{}}: never open, import or \texttt{os.\allowbreak{}path.\allowbreak{}join(\allowbreak{}.\allowbreak{}.\allowbreak{},\allowbreak{} "references",\allowbreak{} .\allowbreak{}.\allowbreak{}.\allowbreak{})} it from code that runs under \texttt{reproduce.\allowbreak{}sh}; copy the files you need into \texttt{submission/\allowbreak{}data/\allowbreak{}} (via \texttt{scripts/\allowbreak{}download\_\allowbreak{}resources.\allowbreak{}sh}, refetching from the same URL) or vendor them under \texttt{src/\allowbreak{}}. Every borrowed block starts with a marker line \texttt{\# ref:\allowbreak{} REF-\allowbreak{}003 path/\allowbreak{}in/\allowbreak{}repo.\allowbreak{}py:\allowbreak{}L40-\allowbreak{}L88} (same idea with \texttt{/\allowbreak{}/\allowbreak{}} in other languages), and \texttt{implementation\_\allowbreak{}map.\allowbreak{}md} gets a \texttt{\#\# References} table: \texttt{REF |\allowbreak{} used in I-\allowbreak{}ids |\allowbreak{} how (\allowbreak{}copied|\allowbreak{}adapted|\allowbreak{}interface|\allowbreak{}config) |\allowbreak{} note}. Borrowed defaults never override the plan's invariants.
\end{skillpanel}

\begin{skillpanel}{promptrev}{Implement Reviewer}
\promptsecfirst{Stage implement, kind build / build-incremental}
Upstream: \texttt{plan.\allowbreak{}md} / \texttt{.\allowbreak{}impl/\allowbreak{}plan\_\allowbreak{}index.\allowbreak{}json}. Review the read-only copy named in the capsule (\texttt{.\allowbreak{}impl/\allowbreak{}review/\allowbreak{}\textless{}sid\textgreater{}/\allowbreak{}repo}), not the live \texttt{submission/\allowbreak{}}.

For each required P: open the file at \texttt{Path:\allowbreak{}:\allowbreak{}Symbol} and \textbf{read the code}. The execution fields in \texttt{experiment-\allowbreak{}exec.\allowbreak{}yaml} are part of the design under review: \texttt{max\_\allowbreak{}parallel}, \texttt{exclusive} and the declared peaks decide how densely the reproduction packs the hardware, so they need a stated basis (code size, batch, what each script loads, the HARDWARE block) rather than a number copied from nowhere; measurements come later in qualify. Trace the \texttt{call\_\allowbreak{}path} yourself from \texttt{scripts/\allowbreak{}run\_\allowbreak{}schedule.\allowbreak{}py} $\rightarrow$ the entry script in \texttt{experiment\_\allowbreak{}schedule.\allowbreak{}json} $\rightarrow$ down to the symbol; WIRED only if the production path calls it. Compare the implementation to the P invariant: different optimizer, activation, sampling, boundary condition, loss term, data split, or default hyperparameter = \texttt{SUBSTITUTED} (critical unless an ambiguity note justifies it). Check config defaults equal the map's claims.

\texttt{build-\allowbreak{}incremental}: the base commit was already reviewed; the capsule lists the impacted P entries. Review those fully; for the rest you may write \texttt{status:\allowbreak{} OK,\allowbreak{} note:\allowbreak{} carried\_\allowbreak{}over} but spot-check at least 3 (or 10\%) by reading code. If an entry file (reproduce.sh, run\_schedule.py, schedule) changed, re-verify every call\_path. Delivery-contract issues (Dockerfile, README, lockfiles) are \texttt{minor}.

\promptsec{Stage implement, kind qualify}
Upstream: the reviewed build commit + \texttt{.\allowbreak{}impl/\allowbreak{}registry.\allowbreak{}json}. Every P0 experiment must have an accepted run (\texttt{reviewed:\allowbreak{} true}, not dirty, status completed); a missing one is \texttt{critical}. Lower priorities are not required, but the exit is stronger the more of them ran: note in the summary how far down the priority list qualification got and whether the remaining budget was used. For each experiment with an accepted run: its \texttt{run.\allowbreak{}log} contains \texttt{[impl] start} and \texttt{[impl] done} produced by the code (grep the source for the \texttt{ImplRun} calls); \texttt{metrics.\allowbreak{}json} values are plausible for the qualification scale and match what \texttt{\#\# Experiments} claims; \texttt{expected\_\allowbreak{}artifacts} are present. Confirm the qualification config only shortens duration (steps/seeds/samples) and keeps shape (batch, model size, sequence length, precision) equal to full, as the registry's \texttt{shape\_\allowbreak{}equivalent} claims; if not, that is a finding. Check every \texttt{\#\# Experiment actions} row against the EVIDENCE block in your capsule: a REUSE whose reason ignores changed code paths, changed configs or newer dependency runs is a finding (major).

\texttt{experiment-\allowbreak{}exec.\allowbreak{}yaml} is under review too: \texttt{full\_\allowbreak{}command} honours the plan's \texttt{protocol} (iterations, seeds, evaluation set; a command that silently shrinks the paper's scale is \texttt{critical}); \texttt{depends\_\allowbreak{}on} names every experiment whose output the code actually reads (a missing real dependency is \texttt{critical}, because the schedule may start it too early); declared \texttt{peak\_\allowbreak{}gpu\_\allowbreak{}gb} / \texttt{peak\_\allowbreak{}ram\_\allowbreak{}gb} are not below the measured peaks in \texttt{.\allowbreak{}impl/\allowbreak{}runs/\allowbreak{}*.\allowbreak{}json} and \texttt{est\_\allowbreak{}minutes\_\allowbreak{}full} agrees with the measured durations scaled to the full command (far off is \texttt{major}); runs are qualify-scale, so a declaration well above a measurement is fine when \texttt{implementation\_\allowbreak{}map.\allowbreak{}md} states the scale-up, and \texttt{peak\_\allowbreak{}gpu\_\allowbreak{}gb:\allowbreak{} null} means the GPU could not be measured (no evidence either way); \texttt{max\_\allowbreak{}parallel} and every \texttt{exclusive:\allowbreak{} true} rest on a measurement, not a guess.
\end{skillpanel}

\begin{skillpanel}{promptprod}{Result Producer}
\promptsecfirst{Your job}
Inputs: \texttt{implementation\_\allowbreak{}map.\allowbreak{}md} (\texttt{\#\# Experiments}), \texttt{.\allowbreak{}impl/\allowbreak{}runs/\allowbreak{}*.\allowbreak{}json} (only \texttt{reviewed:\allowbreak{} true} runs count), their \texttt{output/\allowbreak{}metrics.\allowbreak{}json}, and the S-R rows in \texttt{.\allowbreak{}impl/\allowbreak{}obligations.\allowbreak{}json}.

Output: \texttt{result.\allowbreak{}md} only (you are read-only; \texttt{submission/\allowbreak{}} belongs to the implementer). One table \texttt{ID |\allowbreak{} Spec |\allowbreak{} Run |\allowbreak{} Artifact |\allowbreak{} Value |\allowbreak{} Paper value |\allowbreak{} Verdict}---\texttt{ID} is \texttt{R-\allowbreak{}001}, \texttt{R-\allowbreak{}002}, \ldots{} (your own row ids); \texttt{Spec} holds the \texttt{S-\allowbreak{}R-\allowbreak{}*} obligation ids (list every one the row addresses)---followed by a \texttt{\#\# Deviations} section listing every entry of \texttt{.\allowbreak{}impl/\allowbreak{}deviations.\allowbreak{}jsonl} with one line on how it affects the results. \texttt{Verdict} $\in$ \texttt{SUPPORTED |\allowbreak{} PARTIAL |\allowbreak{} CONTRADICTED |\allowbreak{} NOT\_\allowbreak{}RUN}. \texttt{Value} must be read from \texttt{Artifact}; never type a number from memory or from a log banner. Finish with \texttt{submit(\allowbreak{})} or stop.

\promptsec{Scope of claims}
In this task the runs you see are qualification-scale. A qualification run can support ``pipeline runs end-to-end and produces the artifact with the right shape/name'' and, if the config is close enough, a trend; it cannot support ``the paper's number is reproduced''. Use \texttt{PARTIAL} with a note of the scale, or \texttt{NOT\_\allowbreak{}RUN} when no accepted run exists. Never rerun experiments; never edit code or \texttt{generate\_\allowbreak{}summary.\allowbreak{}py} (if the summary is wrong, say so in a blocker). If a result contradicts the plan or reveals a code bug, attach a blocker (\texttt{target\_\allowbreak{}stage:\allowbreak{} "implement"} or \texttt{"plan"}) with the artifact as evidence.

\promptsec{Blacklist and outside sources}
List every REF the implementation borrowed from (with its license from \texttt{references/\allowbreak{}REF-\allowbreak{}NNN/\allowbreak{}ref.\allowbreak{}json}) in a short \texttt{\#\# References} section of \texttt{result.\allowbreak{}md}.
\end{skillpanel}

\begin{skillpanel}{promptrev}{Result Reviewer}
\promptsecfirst{Stage result}
Upstream: \texttt{implementation\_\allowbreak{}map.\allowbreak{}md} + \texttt{.\allowbreak{}impl/\allowbreak{}runs/\allowbreak{}*.\allowbreak{}json}. For each S-R: an R row exists; open the \texttt{Artifact} and confirm \texttt{Value} is in it; \texttt{Verdict} matches the numbers and the run scale (qualification-scale runs cannot be SUPPORTED for paper numbers); \texttt{generate\_\allowbreak{}summary.\allowbreak{}py} reads \texttt{results/\allowbreak{}} and uses the paper's table/figure names. Findings about code go to \texttt{owner\_\allowbreak{}stage:\allowbreak{} implement}; about the plan to \texttt{plan}.
\end{skillpanel}

\begin{skillpanel}{promptrev}{Reviewer Instructions Shared Across Stages}
\promptsecfirst{Your job}
You review exactly one submission (the capsule names it). You are read-only; \texttt{bash} is for running tests and small scripts only. End your turn with \texttt{review(\allowbreak{}\{\allowbreak{}verdict,\allowbreak{} coverage,\allowbreak{} findings,\allowbreak{} blocker\_\allowbreak{}decisions,\allowbreak{} closed\_\allowbreak{}finding\_\allowbreak{}ids,\allowbreak{} summary\})}. Nothing else ends it.
\begin{itemize}[leftmargin=1.25em,itemsep=1.5pt,topsep=2pt,parsep=0pt]
  \item \texttt{verdict}: \texttt{PASS} if every required id is OK/WIRED and no critical or major finding is open; \texttt{PASS\_\allowbreak{}WITH\_\allowbreak{}ISSUES} if only majors/minors remain open (they are registered as deviations and disclosed); \texttt{REJECT} if any critical is open or a blocker is accepted.
  \item \texttt{coverage}: one row per id in REQUIRED\_IDS: \texttt{\{\allowbreak{}id,\allowbreak{} location,\allowbreak{} call\_\allowbreak{}path,\allowbreak{} status,\allowbreak{} note\}}; status $\in$ \texttt{OK |\allowbreak{} WIRED |\allowbreak{} NOT\_\allowbreak{}WIRED |\allowbreak{} MISSING |\allowbreak{} SUBSTITUTED |\allowbreak{} INCONSISTENT}.
  \item \texttt{findings}: \texttt{\{\allowbreak{}id:\allowbreak{} "F-\allowbreak{}1",\allowbreak{} severity:\allowbreak{} critical|\allowbreak{}major|\allowbreak{}minor,\allowbreak{} obligation,\allowbreak{} location,\allowbreak{} status,\allowbreak{} call\_\allowbreak{}path,\allowbreak{} required\_\allowbreak{}fix,\allowbreak{} owner\_\allowbreak{}stage\}}. \texttt{owner\_\allowbreak{}stage} is the stage that must fix it (may be upstream).
  \item \texttt{blocker\_\allowbreak{}decisions}: one \texttt{\{\allowbreak{}idx,\allowbreak{} decision:\allowbreak{} accepted|\allowbreak{}rejected,\allowbreak{} reason\}} per producer blocker.
  \item \texttt{closed\_\allowbreak{}finding\_\allowbreak{}ids}: previous findings you verified as fixed.
\end{itemize}
Cite paths and line ranges. A finding without a location is not a finding. Findings already listed under REGISTERED DEVIATIONS in your capsule are not re-raised; reference them instead. You may execute code you review: copy the read-only repo to your scratch dir (\texttt{.\allowbreak{}impl/\allowbreak{}scratch/\allowbreak{}\textless{}session\textgreater{}/\allowbreak{}}), build a venv there and run tests or a qualification command. Never run anything inside the live \texttt{submission/\allowbreak{}}.

\promptsec{Three dimensions, one upstream}
Compare this stage's artifact only with its direct upstream (the capsule lists both). For each required id ask:
\begin{enumerate}[leftmargin=1.5em,itemsep=1.5pt,topsep=2pt,parsep=0pt]
  \item \textbf{Existence}---is there an entry / symbol / run / row for it?
  \item \textbf{Correctness}---does the content satisfy the upstream invariant literally (formula, optimizer, boundary condition, sampling grid, data split, hyperparameter defaults, number of seeds)?
  \item \textbf{Consistency}---do the artifact, the code/config defaults and the declared mapping agree with each other?
\end{enumerate}
Severity: an \texttt{exact} invariant violated, a required item missing, or NOT\_WIRED $\rightarrow$ \texttt{critical}; a \texttt{minimum/\allowbreak{}range} shortfall or a mismatch between map and code $\rightarrow$ \texttt{major}; naming/format $\rightarrow$ \texttt{minor}.

\promptsec{Deciding blockers}
Accept when: the upstream entry contradicts its own upstream invariant; the entry is not executable here and no offline path was given; a required item is missing upstream; the \texttt{tried} field shows a real attempt. Reject when: it is solvable inside the producer's own scope (missing config, download script, typo); it is a time/budget problem (the schedule scales); there is no evidence; a qualification-scale number merely disagrees with the paper. An accepted blocker or a critical finding with an upstream \texttt{owner\_\allowbreak{}stage} makes the verdict REJECT and sends the run back to that stage.

\promptsec{Blacklist and outside sources}
You check it as well: no blacklisted identity in the artifact or the code you review; every \texttt{\# ref:\allowbreak{}} marker names a REF that exists under \texttt{references/\allowbreak{}} and the cited path exists in that REF; \texttt{\#\# References} rows agree with the markers; a REF's \texttt{ref.\allowbreak{}json} has \texttt{provenance:\allowbreak{} control-\allowbreak{}plane} (an \texttt{agent-\allowbreak{}reported} provenance is a \texttt{major} finding unless the summary shows the URL and commit). A borrowed block that overrides a plan invariant is \texttt{critical}. For the plan stage the REF check is deliberately shallow: \texttt{ref.\allowbreak{}json} exists and was written by the control plane, the paths cited in \texttt{summary.\allowbreak{}md} exist under the REF, and no blacklisted identity appears in the REF or in \texttt{plan.\allowbreak{}md}; you do not check licenses.
\end{skillpanel}

\endgroup

\FloatBarrier

%% file: tables/cost_time.tex
\begin{table}[!htbp]
\centering
\small
\renewcommand{\arraystretch}{0.95}
\caption{Rollout and reproduction cost per run, over the 40 graded runs. Rollout
h: wall-clock duration of the agent stage; Tok: tokens processed (input, cache read and
output) by all agents; Cost: rollout cost at published list prices; Repro h:
wall-clock duration of the reproduction stage in the fresh environment. Bold: the
reproduction reached the 24\,h execution limit (logged as 24.1\,h wall-clock
including container teardown and artifact collection grace time). --: no reproduction stage (the rollout
aborted before delivery; Table~\ref{tab:main} grades its latest committed snapshot under the full judge, with Code Execution and Result Analysis leaves scoring zero). Mean rows are per paper (counting -- as 0).}
\label{tab:cost}
\setlength{\tabcolsep}{3.5pt}
\begin{tabular*}{\textwidth}{@{\extracolsep{\fill}}l cccc cccc@{}}
\toprule
 & \multicolumn{4}{c}{\textbf{GLM-5}} & \multicolumn{4}{c}{\textbf{DeepSeek-V4-Flash}} \\
\cmidrule(lr){2-5}\cmidrule(l){6-9}
Paper & Rollout h & Tok (M) & Cost \$ & Repro h & Rollout h & Tok (M) & Cost \$ & Repro h \\
\midrule
adaptive-pruning           & 8.9 & 106 & 38.9 & 14.3 & 5.5 & 148 & 4.7 & \textbf{24.1} \\
all-in-one                 & 17.9 & 100 & 33.6 & \textbf{24.1} & 16.0 & 133 & 4.3 & \textbf{24.1} \\
bam                        & 9.5 & 99 & 38.8 & 6.5 & 8.2 & 129 & 4.0 & 2.2 \\
bbox                       & 19.6 & 100 & 41.8 & \textbf{24.1} & 4.6 & 107 & 3.2 & \textbf{24.1} \\
bridging-data-gaps         & 12.2 & 106 & 37.0 & -- & 16.0 & 199 & 6.1 & 8.8 \\
fre                        & 9.1 & 155 & 53.6 & 0.8 & 8.7 & 169 & 4.6 & \textbf{24.1} \\
ftrl                       & 13.9 & 159 & 53.6 & \textbf{24.1} & 12.7 & 162 & 4.4 & \textbf{24.1} \\
lbcs                       & 5.9 & 60 & 21.1 & 19.9 & 8.2 & 134 & 4.5 & 5.1 \\
lca-on-the-line            & 23.7 & 85 & 30.9 & 5.5 & 5.1 & 90 & 2.9 & 11.8 \\
mechanistic-understanding  & 14.3 & 105 & 35.6 & 2.6 & 23.6 & 164 & 5.3 & 1.7 \\
pinn                       & 9.1 & 82 & 29.1 & 23.5 & 23.6 & 196 & 6.2 & 0.7 \\
rice                       & 13.5 & 97 & 34.4 & 16.2 & 8.0 & 126 & 4.0 & 13.8 \\
robust-clip                & 16.8 & 162 & 59.1 & \textbf{24.1} & 21.0 & 169 & 4.9 & 3.9 \\
sample-specific-masks      & 17.8 & 113 & 38.9 & \textbf{24.1} & 9.0 & 137 & 4.4 & 7.5 \\
sapg                       & 8.9 & 102 & 35.6 & 24.0 & 4.8 & 102 & 3.6 & 7.7 \\
sequential-neural          & 19.1 & 179 & 63.6 & \textbf{24.1} & 10.9 & 113 & 3.5 & \textbf{24.1} \\
stay-on-topic              & 14.3 & 145 & 50.8 & \textbf{24.1} & 5.5 & 110 & 3.7 & \textbf{24.1} \\
stochastic-interpolants    & 13.9 & 64 & 26.2 & 23.9 & 23.6 & 144 & 4.3 & \textbf{24.1} \\
test-time-model-adaptation & 20.0 & 156 & 58.2 & 0.8 & 10.2 & 168 & 5.2 & 16.9 \\
what-will-my-model-forget  & 19.2 & 114 & 45.8 & 5.9 & 9.3 & 97 & 2.9 & 24.0 \\
\midrule
\textbf{Mean} & 14.4 & 115 & 41.3 & 15.6 & 11.7 & 140 & 4.3 & 14.8 \\
\bottomrule
\end{tabular*}
\end{table}

%% file: references.bib
@inproceedings{seo2026paper2code,
  title={{Paper2Code}: Automating Code Generation from Scientific Papers in Machine Learning},
  author={Seo, Minju and Baek, Jinheon and Lee, Seongyun and Hwang, Sung Ju},
  booktitle={The Fourteenth International Conference on Learning Representations},
  year={2026},
  url={https://openreview.net/forum?id=3DcaUTjdKc}
}

@article{lu2026automation,
  title     = {Towards end-to-end automation of {AI} research},
  author    = {Lu, Chris and Lu, Cong and Lange, Robert Tjarko and Yamada, Yutaro and Hu, Shengran and Foerster, Jakob and Ha, David and Clune, Jeff},
  journal   = {Nature},
  volume    = {651},
  number    = {8107},
  year      = {2026},
  pages     = {914--919},
  doi       = {10.1038/s41586-026-10265-5},
}

@article{yamada2025aiscientistv2,
  title={The {AI} Scientist-v2: Workshop-Level Automated Scientific Discovery via Agentic Tree Search},
  author={Yamada, Yutaro and Lange, Robert Tjarko and Lu, Cong and Hu, Shengran and Lu, Chris and Foerster, Jakob and Clune, Jeff and Ha, David},
  journal={arXiv preprint arXiv:2504.08066},
  year={2025}
}

@inproceedings{schmidgall2025agentlaboratory,
    title = "Agent Laboratory: Using {LLM} Agents as Research Assistants",
    author = "Schmidgall, Samuel  and
      Su, Yusheng  and
      Wang, Ze  and
      Sun, Ximeng  and
      Wu, Jialian  and
      Yu, Xiaodong  and
      Liu, Jiang  and
      Moor, Michael  and
      Liu, Zicheng  and
      Barsoum, Emad",
    editor = "Christodoulopoulos, Christos  and
      Chakraborty, Tanmoy  and
      Rose, Carolyn  and
      Peng, Violet",
    booktitle = "Findings of the Association for Computational Linguistics: EMNLP 2025",
    month = nov,
    year = "2025",
    address = "Suzhou, China",
    publisher = "Association for Computational Linguistics",
    url = "https://aclanthology.org/2025.findings-emnlp.320/",
    doi = "10.18653/v1/2025.findings-emnlp.320",
    pages = "5977--6043",
    ISBN = "979-8-89176-335-7",
}

@article{zhu2025aiscientistsfailstrong,
      title={{AI} Scientists Fail Without Strong Implementation Capability}, 
      author={Minjun Zhu and Qiujie Xie and Yixuan Weng and Jian Wu and Zhen Lin and Linyi Yang and Yue Zhang},
      year={2025},
  journal = {arXiv preprint arXiv:2506.01372},
      url={https://arxiv.org/abs/2506.01372}, 
}

@inproceedings{autop2c2025,
  title     = {{AutoP2C}: An {LLM}-Based Agent Framework for Code Repository Generation from Multimodal Content in Machine Learning Papers},
  author    = {Lin, Zijie and Shen, Yiqing and Cai, Qilin and Sun, He and Zhou, Jinrui and Xiao, Mingjun},
  booktitle = {2026 IEEE International Conference on Acoustics, Speech and Signal Processing (ICASSP)},
  year      = {2026},
  pages     = {12487--12491},
  publisher = {IEEE},
  doi       = {10.1109/icassp55912.2026.11462205},
}

@inproceedings{autoreproduce2025,
  title     = {{AutoReproduce}: Automatic {AI} Experiment Reproduction with Paper Lineage},
  author    = {Zhao, Xuanle and Sang, Zilin and Li, Yuxuan and Shi, Qi and Zhao, Weilun and Wang, Shuo and Zhang, Duzhen and Han, Xu and Liu, Zhiyuan and Sun, Maosong},
  booktitle = {Proceedings of the 64th Annual Meeting of the {A}ssociation for {C}omputational {L}inguistics (Volume 1: Long Papers)},
  year      = {2026},
  pages     = {21920--21942},
  publisher = {Association for Computational Linguistics},
  doi       = {10.18653/v1/2026.acl-long.1001},
  url       = {https://aclanthology.org/2026.acl-long.1001/},
}

@article{li2025deepcode,
  title={{DeepCode}: Open Agentic Coding},
  author={Li, Zongwei and Li, Zhonghang and Guo, Zirui and Ren, Xubin and Huang, Chao},
  journal={arXiv preprint arXiv:2512.07921},
  year={2025}
}

@inproceedings{zhou2025repro,
  title     = {{SigRef}: Verification-Driven Reflection for Faithful Paper-to-Code Development},
  author    = {Zhou, Mingyang and Yao, Quanming and Du, Lun and Wei, Lanning and Zheng, Da},
  booktitle = {Pattern Recognition. {ICPR} 2026},
  series    = {Lecture Notes in Computer Science},
  volume    = {16825},
  year      = {2027},
  pages     = {172--186},
  publisher = {Springer Nature Switzerland},
  doi       = {10.1007/978-3-032-31930-2_12},
}

@inproceedings{chen2025deepreproducer,
  title     = {{Deep-Reproducer}: From Paper Understanding to Code Generation},
  author    = {Chen, Pengcheng and Yan, Ning and Zhao, Zihan and Lin, Yixiao and Chen, Huaibo and Hu, Yue and Bai, Qinbo and Li, Xiang and Mortazavi, Masood S.},
  booktitle = {NeurIPS 2025 Fourth Workshop on Deep Learning for Code},
  year      = {2025},
  url       = {https://openreview.net/forum?id=zw2DpSxnXn},
}

@inproceedings{hiras2026,
  title     = {{HiRAS}: A Hierarchical Multi-Agent Framework for Paper-to-Code Generation and Execution},
  author    = {Hong, Hanhua and Li, Yizhi and Chen, Jiaoyan and Ananiadou, Sophia and Li, Xiaoli and Kim, Jung-jae and Lin, Chenghua},
  booktitle = {Findings of the {A}ssociation for {C}omputational {L}inguistics: {ACL} 2026},
  year      = {2026},
  pages     = {7624--7652},
  publisher = {Association for Computational Linguistics},
  doi       = {10.18653/v1/2026.findings-acl.377},
  url       = {https://aclanthology.org/2026.findings-acl.377/},
}

@inproceedings{
    starace2025paperbench,
    title={{PaperBench}: Evaluating {AI}{\textquoteright}s Ability to Replicate {AI} Research},
    author={Giulio Starace and Oliver Jaffe and Dane Sherburn and James Aung and Jun Shern Chan and Leon Maksin and Rachel Dias and Evan Mays and Benjamin Kinsella and Wyatt Thompson and Johannes Heidecke and Amelia Glaese and Tejal Patwardhan},
    booktitle={Forty-second International Conference on Machine Learning},
    year={2025},
    url={https://openreview.net/forum?id=xF5PuTLPbn}
}

@article{whynotscientists2026,
  title   = {Why {LLM}s Aren't Scientists Yet: Lessons from Four Autonomous Research Attempts},
  author  = {Trehan, Dhruv and Chopra, Paras},
  journal = {arXiv preprint arXiv:2601.03315},
  year    = {2026},
  url     = {https://arxiv.org/abs/2601.03315},
}

@inproceedings{cemri2025multiagentfail,
  title     = {Why Do Multi-Agent {LLM} Systems Fail?},
  author    = {Cemri, Mert and Pan, Melissa Z. and Yang, Shuyi and Agrawal, Lakshya A. and Chopra, Bhavya and Tiwari, Rishabh and Keutzer, Kurt and Parameswaran, Aditya and Klein, Dan and Ramchandran, Kannan and Zaharia, Matei and Gonzalez, Joseph E. and Stoica, Ion},
  booktitle = {Advances in Neural Information Processing Systems},
  volume    = {38},
  year      = {2025},
  pages     = {135676--135729},
  publisher = {Curran Associates, Inc.},
  doi       = {10.52202/085713-4082},
}

@inproceedings{aiscientistengineering2026,
  title     = {Toward Autonomous Long-Horizon Engineering for {ML} Research},
  author    = {Chen, Guoxin and Chen, Jie and Chen, Lei and Zhao, Jiale and Meng, Fanzhe and Zhao, Wayne Xin and Song, Ruihua and Chen, Cheng and Wen, Ji-Rong and Jia, Kai},
  booktitle = {Findings of the {A}ssociation for {C}omputational {L}inguistics: {EMNLP} 2026},
  year      = {2026},
  pages     = {47116--47131},
  publisher = {Association for Computational Linguistics},
  doi       = {10.18653/2026.findings-emnlp.2200},
  url       = {https://aclanthology.org/2026.findings-emnlp.2200/},
}

@misc{openai2026gpt54,
  title   = {Introducing {GPT}-5.4},
  author  = {{OpenAI}},
  year    = {2026},
  url     = {https://openai.com/index/introducing-gpt-5-4/},
}

@article{zeng2026glm,
  title={{GLM-5}: From Vibe Coding to Agentic Engineering},
  author={Zeng, Aohan and Lv, Xin and Hou, Zhenyu and Du, Zhengxiao and Zheng, Qinkai and Chen, Bin and Yin, Da and Ge, Chendi and Huang, Chenghua and Xie, Chengxing and others},
  journal={arXiv preprint arXiv:2602.15763},
  year={2026}
}

@misc{deepseekai2026v4release,
  author       = {{DeepSeek-AI}},
  title        = {{DeepSeek V4 Preview Release}},
  year         = {2026},
  howpublished = {\url{https://api-docs.deepseek.com/news/news260424}},
  note         = {Accessed: 2026-05-27}
}

@inproceedings{reproagent2026,
  title     = {{ReproAgent}: Contract-Guided Paper-to-Code Reproduction},
  author    = {Hu, Xue and Pan, Zewei and Wang, Zhongyuan and Liu, Zhou and Su, Zeli and Zhang, Wentao},
  booktitle = {Findings of the {A}ssociation for {C}omputational {L}inguistics: {EMNLP} 2026},
  year      = {2026},
  pages     = {3681--3702},
  publisher = {Association for Computational Linguistics},
  doi       = {10.18653/2026.findings-emnlp.175},
  url       = {https://aclanthology.org/2026.findings-emnlp.175/},
}

@inproceedings{deeprepro2026,
  title     = {{DeepRepro}: State-Aware Subplanning for Paper-to-Code Reproduction in Evolving Repositories},
  author    = {Song, Hongru and Zhang, Ruqing and Guo, Jiafeng and Cheng, Xueqi and de Rijke, Maarten},
  booktitle = {Proceedings of the 35th ACM International Conference on Information and Knowledge Management (CIKM), Demo Track},
  year      = {2026},
  pages     = {9250--9254},
  publisher = {ACM},
  doi       = {10.1145/3799682.3840257},
}

@article{milo2026,
  title   = {{MILO}: Automated Harness Discovery via Orchestrated Multi-Agent Evolution},
  author  = {Jana, Prithwish and Goswami, Mononito and Liu, Hao and Li, Xinyu and Huang, Langlin and Huang, Zhehui and Huang, Zhishen and Bl{\"o}baum, Patrick and Deoras, Anoop and Jain, Purak and Kanakaris, Nikos and Genc, Sahika},
  journal = {arXiv preprint arXiv:2609.38349},
  year    = {2026}
}

@article{zhang2025darwin,
  title     = {Darwin G{\"o}del Machine: Open-Ended Evolution of Self-Improving Agents},
  author    = {Zhang, Jenny and Hu, Shengran and Lu, Cong and Lange, Robert and Clune, Jeff},
  journal   = {SuperIntelligence - Robotics - Safety \& Alignment},
  volume    = {2},
  number    = {3},
  year      = {2025},
  doi       = {10.70777/si.v2i3.15063},
}

@inproceedings{cai2024bam,
  title     = {Batch and Match: Black-Box Variational Inference with a Score-Based Divergence},
  author    = {Cai, Diana and Modi, Chirag and Pillaud-Vivien, Loucas and Margossian, Charles C. and Gower, Robert M. and Blei, David M. and Saul, Lawrence K.},
  booktitle = {Proceedings of the 41st International Conference on Machine Learning},
  year      = {2024}
}

@article{autoresearchclaw2026,
  title = {{AutoResearchClaw: Self-Reinforcing Autonomous Research with Human-AI Collaboration}},
  author = {Liu, Jiaqi and Qiu, Shi and Li, Mairui and Li, Bingzhou and Ji, Haonian and Han, Siwei and Ye, Xinyu and Xia, Peng and Dong, Zihan and Chen, Meng and Zhang, Congyu and Zhang, Letian and Chen, Guiming and Tu, Haoqin and Yang, Xinyu and Feng, Lu and Zhao, Xujiang and Chen, Haifeng and Zhou, Jiawei and Wang, Xiao and Zhang, Weitong and Zhu, Hongtu and Li, Yun and Mei, Jieru and Fei, Hongliang and Zhang, Jiaheng and Li, Linjie and Zhang, Linjun and Zhou, Yuyin and Wang, Sheng and Xiong, Caiming and Zou, James and Zheng, Zeyu and Xie, Cihang and Ding, Mingyu and Yao, Huaxiu},
  year = {2026},
  journal = {arXiv preprint arXiv:2605.20025},
}

@article{scientistone2026,
  title = {{ScientistOne: Towards Human-Level Autonomous Research via Chain-of-Evidence}},
  author = {Meng, Rui and Mishra, Bhavana Dalvi and Chen, Jiefeng and Li, Chun-Liang and Goyal, Palash and Parmar, Mihir and Song, Yiwen and Song, Yale and Sinha, Rajarishi and Ranganathan, Parthasarathy and Gokturk, Burak and Yoon, Jinsung and Pfister, Tomas},
  year = {2026},
  journal = {arXiv preprint arXiv:2605.26340},
}

@article{paperreplication2026,
  title = {{Coding-agents Can Replicate Scientific Machine Learning Papers}},
  author = {Hans, Atharva and Bilionis, Ilias},
  year = {2026},
  journal = {arXiv preprint arXiv:2607.02134},
}

@article{papercompiler2026,
  title = {{PaperCompiler: Faithful Paper-to-Code Generation via Repository-Level Specification Compilation}},
  author = {Liu, Yunhao and Pham, Hong Phuc and Yoon, Jaehong},
  year = {2026},
  journal = {arXiv preprint arXiv:2609.02272},
}

@inproceedings{scicoqa2026,
  title     = {{SciCoQA}: Quality Assurance for Scientific Paper--Code Alignment},
  author    = {Baumg{\"a}rtner, Tim and Gurevych, Iryna},
  booktitle = {Proceedings of the 64th Annual Meeting of the {A}ssociation for {C}omputational {L}inguistics (Volume 1: Long Papers)},
  year      = {2026},
  pages     = {38740--38770},
  publisher = {Association for Computational Linguistics},
  doi       = {10.18653/v1/2026.acl-long.1795},
  url       = {https://aclanthology.org/2026.acl-long.1795/},
}

@article{kon2025curie,
  title = {{Curie: Toward Rigorous and Automated Scientific Experimentation with AI Agents}},
  author = {Kon, Patrick Tser Jern and Liu, Jiachen and Ding, Qiuyi and Qiu, Yiming and Yang, Zhenning and Huang, Yibo and Srinivasa, Jayanth and Lee, Myungjin and Chowdhury, Mosharaf and Chen, Ang},
  year = {2025},
  journal = {arXiv preprint arXiv:2502.16069},
}

@inproceedings{hua2025researchcodebench,
  title     = {{ResearchCodeBench}: Benchmarking {LLMs} on Implementing Novel Machine Learning Research Code},
  author    = {Hua, Tianyu and Hua, Harper and Xiang, Violet and Klieger, Benjamin and Truong, Sang T. and Liang, Weixin and Sun, Fan-Yun and Haber, Nick},
  booktitle = {Advances in Neural Information Processing Systems},
  volume    = {38},
  year      = {2025},
  pages     = {155614--155648},
  publisher = {Curran Associates, Inc.},
  doi       = {10.52202/085713-4687},
}

@article{rda2025,
  title = {{R\&D-Agent: An LLM-Agent Framework Towards Autonomous Data Science}},
  author = {Yang, Xu and Yang, Xiao and Fang, Shikai and Zhang, Yifei and Wang, Jian and Xian, Bowen and Li, Qizheng and Li, Jingyuan and Xu, Minrui and Li, Yuante and Pan, Haoran and Zhang, Yuge and Liu, Weiqing and Shen, Yelong and Chen, Weizhu and Bian, Jiang},
  year = {2025},
  journal = {arXiv preprint arXiv:2505.14738},
}

@article{zhang2026hyperagents,
  title = {{Hyperagents}},
  author = {Zhang, Jenny and Zhao, Bingchen and Yang, Wannan and Foerster, Jakob and Clune, Jeff and Jiang, Minqi and Devlin, Sam and Shavrina, Tatiana},
  year = {2026},
  journal = {arXiv preprint arXiv:2603.19461},
}

@article{lee2026metaharness,
  title = {{Meta-Harness: End-to-End Optimization of Model Harnesses}},
  author = {Lee, Yoonho and Nair, Roshen and Zhang, Qizheng and Lee, Kangwook and Khattab, Omar and Finn, Chelsea},
  year = {2026},
  journal = {arXiv preprint arXiv:2603.28052},
}

@article{karten2026continual,
  title = {{Continual Harness: Online Adaptation for Self-Improving Foundation Agents}},
  author = {Karten, Seth and Zhang, Joel and Upaa Jr, Tersoo and Feng, Ruirong and Li, Wenzhe and Shi, Chengshuai and Jin, Chi and Vodrahalli, Kiran},
  year = {2026},
  journal = {arXiv preprint arXiv:2605.09998},
}

@article{shen2026skilloptlite,
  title = {{SkillOpt-Lite: Better and Faster Agent Self-evolution via One Line of Vibe}},
  author = {Shen, Yifei and Li, Bo and Zhang, Xinjie},
  year = {2026},
  journal = {arXiv preprint arXiv:2607.03451},
}

@article{zhang2026selfharness,
  title = {{Self-Harness: Harnesses That Improve Themselves}},
  author = {Zhang, Hangfan and Zhang, Shao and Li, Kangcong and Zhang, Chen and Chen, Yang and Zhang, Yiqun and Bai, Lei and Hu, Shuyue},
  year = {2026},
  journal = {arXiv preprint arXiv:2606.09498},
}

@inproceedings{wen2025mislead,
  title = {Language Models Learn to Mislead Humans via {RLHF}},
  author = {Wen, Jiaxin and Zhong, Ruiqi and Khan, Akbir and Perez, Ethan and Steinhardt, Jacob and Huang, Minlie and Bowman, Samuel R. and He, He and Feng, Shi},
  booktitle = {International Conference on Learning Representations},
  year = {2025},
}

@article{baker2025monitoring,
  title = {{Monitoring Reasoning Models for Misbehavior and the Risks of Promoting Obfuscation}},
  author = {Baker, Bowen and Huizinga, Joost and Gao, Leo and Dou, Zehao and Guan, Melody Y. and Madry, Aleksander and Zaremba, Wojciech and Pachocki, Jakub and Farhi, David},
  year = {2025},
  journal = {arXiv preprint arXiv:2503.11926},
}

@article{internagent2025,
  title = {{InternAgent: When Agent Becomes the Scientist -- Building Closed-Loop System from Hypothesis to Verification}},
  author = {{InternAgent Team} and Zhang, Bo and Feng, Shiyang and Yan, Xiangchao and Yuan, Jiakang and Ma, Runmin and Hu, Yusong and Yu, Zhiyin and He, Xiaohan and Huang, Songtao and Hou, Shaowei and Nie, Zheng and Wang, Zhilong and Liu, Jinyao and Peng, Tianshuo and Ye, Peng and Zhou, Dongzhan and Zhang, Shufei and Wang, Xiaosong and Zhang, Yilan and Li, Meng and Tu, Zhongying and Yue, Xiangyu and Ouyang, Wanli and Zhou, Bowen and Bai, Lei},
  year = {2025},
  journal = {arXiv preprint arXiv:2505.16938},
}

@inproceedings{tang2025airesearcher,
  title     = {{AI-Researcher}: Autonomous Scientific Innovation},
  author    = {Tang, Jiabin and Xia, Lianghao and Li, Zhonghang and Huang, Chao},
  booktitle = {Advances in Neural Information Processing Systems},
  volume    = {38},
  year      = {2025},
  pages     = {10477--10516},
  publisher = {Curran Associates, Inc.},
  doi       = {10.52202/085713-0320},
}

@article{lyu2026evoscientist,
  title = {{EvoScientist: Towards Multi-Agent Evolving AI Scientists for End-to-End Scientific Discovery}},
  author = {Lyu, Yougang and Zhang, Xi and Yi, Xinhao and Zhao, Yuyue and Guo, Shuyu and Hu, Wenxiang and Piotrowski, Jan and Kaliski, Jakub and Urbani, Jacopo and Meng, Zaiqiao and Zhou, Lun and Yan, Xiaohui},
  year = {2026},
  journal = {arXiv preprint arXiv:2603.08127},
}

@article{veritas2026,
  title = {{VERITAS: Towards a General-Purpose Replication Tool for Scientific Research}},
  author = {Liu, Haokun and Tjiaranata, Filbert Aurelian and Tan, Chenhao},
  year = {2026},
  journal = {arXiv preprint arXiv:2607.02931},
}

@article{repotoskill2026,
  title = {{Repo-To-Skill}: Distilling {GitHub} Repositories Into {AI4AI} Skills},
  author = {Chen, Jianlyu and Hu, Yuyang and Qian, Hongjin and Liu, Jiawei and Wei, Wenqing and Chen, Xiaolong and Lian, Defu and Dou, Zhicheng and Li, Chaozhuo and Ye, Qiwei and Liu, Zheng},
  year = {2026},
  journal = {arXiv preprint arXiv:2609.02749},
}

@article{clear2026,
  title = {{CLEAR}: Structured Specifications for Controllable Long-Horizon {ML} Research Engineering},
  author = {Liu, Xin and Song, Hongru and Zhang, Ruqing and Guo, Jiafeng and Cheng, Xueqi},
  year = {2026},
  journal = {arXiv preprint arXiv:2611.06263},
}
